\documentclass{article}

\PassOptionsToPackage{numbers, compress}{natbib}
\usepackage[final]{neurips_data_2023}

\usepackage[utf8]{inputenc} 
\usepackage[T1]{fontenc}    
\usepackage[colorlinks=true, citecolor=green]{hyperref}       
\usepackage{url}            
\usepackage{booktabs}       
\usepackage{amsfonts}       
\usepackage{nicefrac}       
\usepackage{microtype}      
\usepackage[table]{xcolor} 
\usepackage{times}
\usepackage{epsfig}
\usepackage{graphicx}
\usepackage{amsmath}
\usepackage{amssymb}
\usepackage{adjustbox}
\usepackage{tabularx}
\usepackage{tabularray}
\usepackage{booktabs} 
\usepackage{multirow}
\usepackage{xspace}
\usepackage{caption}
\usepackage{stackengine}
\usepackage{makecell}
\usepackage{array}
\usepackage{utfsym}
\usepackage{bbding}
\usepackage{subfig}
\usepackage{enumerate}
\usepackage[lined,boxed,commentsnumbered, ruled]{algorithm2e}

\makeatletter
\DeclareRobustCommand\onedot{\futurelet\@let@token\@onedot}
\def\@onedot{\ifx\@let@token.\else.\null\fi\xspace}
 
\def\ie{\emph{i.e}\onedot}

\makeatother

\begin{document}

\title{Occ3D: A Large-Scale 3D Occupancy Prediction Benchmark for Autonomous Driving}

%

\author{Xiaoyu Tian$\phantom{}^{1}$\thanks{Authors contributed equally.}\hspace{8pt}
Tao Jiang$\phantom{}^{1,3}  \phantom{}^{*}$ \hspace{6pt}
Longfei Yun$\phantom{}^{1}$ \hspace{6pt}
Yucheng Mao$\phantom{}^{1}$ \hspace{6pt}
Huitong Yang$\phantom{}^{4}$ \hspace{6pt} \vspace{6pt} \\
\textbf{Yue Wang$\phantom{}^{2}$\hspace{6pt}
Yilun Wang$\phantom{}^{1}$\hspace{6pt}
Hang Zhao$\phantom{}^{1,3,4}$\thanks{Corresponding to: hangzhao@mail.tsinghua.edu.cn}
\vspace{6pt}} \\
$\phantom{}^1$IIIS, Tsinghua University\hspace{10pt} \\
$\phantom{}^2$University of Southern California\hspace{10pt} 
$\phantom{}^3$Shanghai AI Lab\hspace{10pt}
$\phantom{}^4$Shanghai Qi Zhi Institute
} 


\onecolumn{%
\renewcommand\onecolumn[1][]{#1}%
\maketitle

\begin{center}
    \centering
    \vspace{-10pt}
    \captionsetup{type=figure}
    \includegraphics[width=1.0\textwidth]{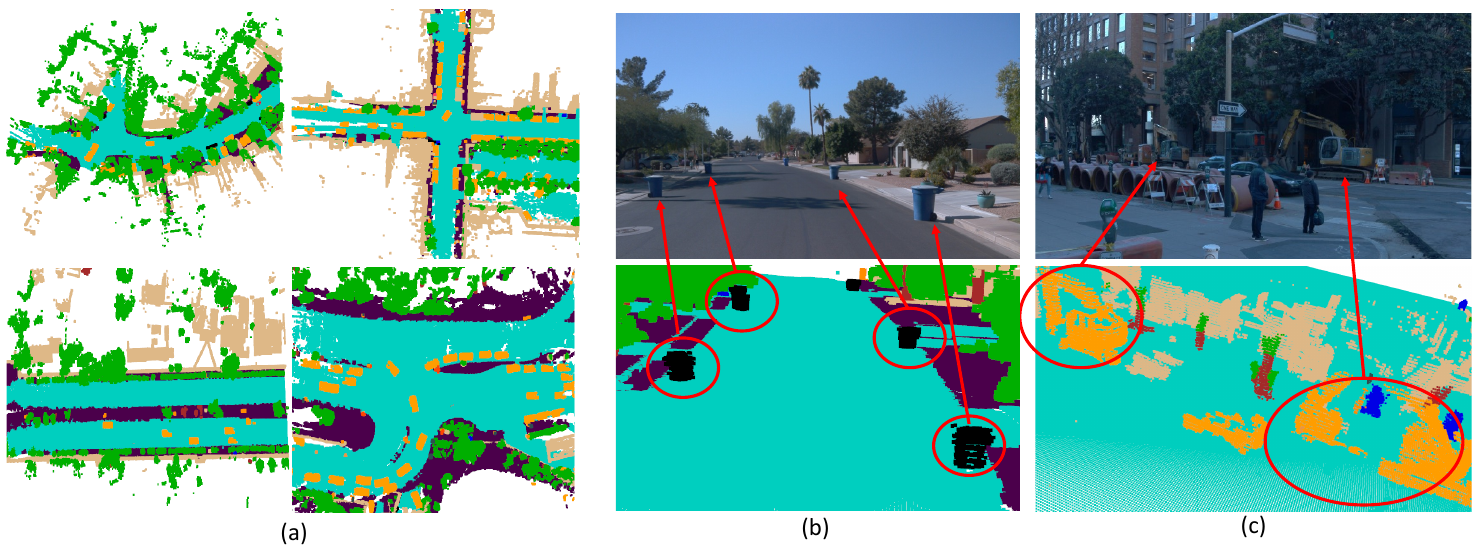}
    \captionof{figure}{\textbf{Our Occ3D dataset demonstrates rich semantic and geometric expressiveness.} (a) Diversity of scenes in the Occ3D dataset; (b) Out-of-vocabulary objects, also known as General Objects (GOs), that cannot be extensively enumerated in the real world; (c) Irregularly-shaped objects that 3D bounding boxes fail to represent their accurate geometry. } 
    \label{fig:teaser}
\end{center}%
}


\begin{abstract}
\vspace{-5pt}
Robotic perception requires the modeling of both 3D geometry and semantics. Existing methods typically focus on estimating 3D bounding boxes, neglecting finer geometric details and struggling to handle general, out-of-vocabulary objects. 3D occupancy prediction, which estimates the detailed occupancy states and semantics of a scene, is an emerging task to overcome these limitations.
To support 3D occupancy prediction, we develop a label generation pipeline that produces dense, visibility-aware labels for any given scene. This pipeline comprises three stages: voxel densification, occlusion reasoning, and image-guided voxel refinement. We establish two benchmarks, derived from the Waymo Open Dataset and the nuScenes Dataset, namely Occ3D-Waymo and Occ3D-nuScenes benchmarks. 
Furthermore, we provide an extensive analysis of the proposed dataset with various baseline models. 
Lastly, we propose a new model, dubbed Coarse-to-Fine Occupancy (CTF-Occ) network, which demonstrates superior performance on the Occ3D benchmarks.
The code, data, and benchmarks are released at \url{https://tsinghua-mars-lab.github.io/Occ3D/}. 
\end{abstract}

\section{Introduction}
\label{sec:intro}


3D perception is a crucial component in vision-based robotic systems like autonomous driving. 
One of the most popular visual perception tasks is 3D object detection, which estimates the 3D locations and dimensions of objects defined in a pre-determined ontology tree~\cite{wang2022detr3d, li2022bevformer}. While the resulting 3D bounding boxes are compact, the level of expressiveness they provide is restricted, as illustrated in Figure \ref{fig:teaser}: (1) 3D bounding box representation erases the geometric details of objects, 
a construction vehicle has a mechanical arm that protrudes from the main body; (2) uncommon categories, like trash cans on the streets, are often ignored and not labeled in the datasets~\cite{caesar2020nuscenes, sun2020scalability} since object categories in the open world cannot be extensively enumerated.

These limitations call for a general and coherent representation that can model the detailed geometry and semantics of objects both within and outside of the ontology tree. 3D Occupancy Prediction, \ie understanding every voxel in the 3D space, is an important task to achieving this goal.
We formalize the 3D occupancy prediction task as follows: a model needs to jointly estimate the \textit{occupancy state} and \textit{semantic label} of every voxel in the scene from images~\cite{behley2019iccv,liao2022kitti,cao2022monoscene}. 
The occupancy state of each voxel can be categorized as \textit{free}, \textit{occupied}, or \textit{unobserved}. 
For occupied voxels, semantic labels are assigned. For objects that are not in the predefined categories, they are labeled as \textit{General Objects (GOs)}. Although GOs are rare, they are essential for perception tasks with safety considerations since they are typically undetected by 3D object detection with predefined categories.

Despite recent advancements in 3D occupancy prediction~\cite{cao2022monoscene,huang2023tri,zhang2023occformer}, there is a notable absence of high-quality datasets together with benchmarks. Constructing such a dataset is challenging due to three major issues: sparsity, occlusion and 3D-2D misalignment. To overcome these hurdles, we create a semi-automatic label generation pipeline that consists of three steps: voxel densification, occlusion reasoning, and image-guided voxel refinement.
Each step within our pipeline is validated through a 3D-2D consistency metric, demonstrating that our proposed label generation pipeline effectively generates dense and visibility-aware annotations.

Building upon the public Waymo Open Dataset~\cite{sun2020scalability}, nuScenes~\cite{caesar2020nuscenes} and Panoptic nuScenes~\cite{nuscenes-lidarseg} Dataset, we produce two benchmarks for our task accordingly, Occ3D-Waymo and Occ3D-nuScenes. Compared to conventional datasets such as SemanticKITTI \cite{behley2019iccv} and KITTI-360 \cite{liao2022kitti}, our Occ3D is the first dataset to offer the surround-view images and high-resolution 3D voxel occupancy representation with the most diverse scenarios. 

A series of recent occupancy prediction models are reproduced and benchmarked on Occ3D.
Additionally, we propose CTF-Occ, a transformer-based \textbf{C}oarse-\textbf{T}o-\textbf{F}ine 3D \textbf{Occ}upancy prediction network. CTF-Occ achieves superior performance by aggregating 2D image features into 3D space via cross-attention in an efficient coarse-to-fine fashion.

The contributions of this work are as follows: (1) We introduce Occ3D, a high-quality 3D occupancy prediction benchmark to facilitate research in this emerging area; (2) We put forward a rigorous automatic label generation pipeline for constructing the Occ3D benchmark, with comprehensive validation of the effectiveness of the pipeline; (3) We benchmark existing model and propose a new CTF-Occ network that achieves superior 3D occupancy prediction performance.

\section{Related Work}
\noindent\textbf{3D detection.}
The goal of 3D object detection is to estimate the locations and dimensions of objects within a predefined ontology.
3D object detection is often performed in LiDAR point clouds~\cite{voxelnet, pointpillar, second, centerpoint, pointnet, pointnet++, ding2019votenet, rukhovich2022fcaf3d}.
More recently, vision-based 3D object detection has gained more attention due to its low cost and rich semantic content~\cite{monodis, fcos3d, wang2022detr3d, li2022bevformer, liu2022petr, philion2020lift, liu2022bevfusion, hu2021fiery, liu2022petrv2, huang2021bevdet, li2022bevdepth, park2022time, lin2022sparse4d}.
Several LiDAR-camera fusion methods are also proposed~\cite{frustumpointnet, chen2022futr3d, liu2022bevfusion}.

\noindent\textbf{3D occupancy prediction.}
A related task of 3D occupancy prediction is Occupancy Grid Mapping (OGM)~\cite{moravec1985high,thrun2002probabilistic,wang2023openoccupancy}, a classical task in mobile robots that aims to generate probabilistic maps from sequential noisy range measurements. 
OGM can be solved within a Bayesian framework, some recent works further combine semantic segmentation with OGM for downstream tasks~\cite{jeon2018traffic,sless2019road,roddick2020predicting}. 
Note that OGM requires range sensors, and also makes the assumption that the scene is static over time. The 3D occupancy prediction task does not have these constraints and can be applied in vision-only robotic systems in dynamic scenes.
Recently, TPVFormer~\cite{huang2023tri} proposes a tri-perspective view method to predict 3D occupancy. However, its output is sparse due to LiDAR supervision. 

\noindent\textbf{Semantic scene completion.}
Another related task is Semantic Scene Completion (SSC)~\cite{armeni2017joint, chang2017matterport3d, dai2017scannet, behley2019dataset, liao2022kitti, yan2021sparse, roldao2020lmscnet,wei2023surroundocc,chen20203d,li2020anisotropic,yan2021sparse,peng2020convolutional,voxformer,occdepth,bev-io}, whose goal is to estimate a dense semantic space from partial observations. SSC differs from 3D occupancy prediction in two ways: (1) SSC focuses on inferring occluded regions given visible parts, while occupancy prediction does not intend to estimate the invisible regions; (2) existing SSC task typically deals with static scenes, whereas occupancy prediction works with dynamic ones.

\begin{table*}[t]
	\scriptsize
	\setlength{\tabcolsep}{0.0035\linewidth}
	\newcommand{\nclass}[1]{\scriptsize{#1}}
	\def\mystrut{\rule{0pt}{1.5\normalbaselineskip}}
	\caption{\textbf{Dataset comparison}. Comparing Occ3D Datasets with other occupancy prediction datasets. Surround  = $\checkmark$ represents surround-view image inputs. C, D, L denote camera, depth and LiDAR.}
	\centering
	\rowcolors[]{3}{black!3}{white}
	\begin{adjustbox}{width=0.99\columnwidth,center}
	\begin{tabular}{l | c c c c c c c c}
		\toprule
		Dataset & Type & Surround & Modality & \# Classes  & \# Sequences & \# Frames  & Volume Size & Resolution (m) \\ 
		\midrule 
        NYUv2 ~\cite{silberman2012indoor} & Indoor &   \usym{2717} & C \& D  & 11  & 464  & 1449 & [240, 240, 14] & - \\  
        ScanNet ~\cite{dai2017scannet}   & Indoor &   \usym{2717} & C \& D  & 11  & 1513 & 1513 & [62, 62, 31] & - \\       

        
        SemanticKITTI~\cite{behley2019iccv} & Outdoor & \usym{2717} & C \& L  & 28  & 22 & 4,3000 & [256, 256, 32] & [0.2, 0.2, 0.2]\\ 
        KITTI-360~\cite{Liao2021ARXIV}  & Outdoor & Fisheye & C \& L  & 19 & 11 & 90,960 & [256, 256, 32] & [0.2, 0.2, 0.2]\\
        
        \midrule 
        \textbf{Occ3D-nuScenes} & Outdoor  & \usym{2713} & C \& L & \nclass{16}+\textit{GO}    & 1000 & 40,000 & [200, 200, 16] & [0.4, 0.4, 0.4]\\ 
        \textbf{Occ3D-Waymo} & Outdoor & \usym{2713} & C \& L   & 14+\textit{GO}  & 1000 & 200,000 & [3200, 3200, 128] & [0.05, 0.05, 0.05]\\
		\toprule
	\end{tabular}
	\end{adjustbox}

	\label{table:datasets}
\end{table*}

\section{Occ3D Dataset}
\subsection{Task Definition}

Given a sequence of sensor inputs, the goal of 3D occupancy prediction is to estimate the state of each voxel in the 3D scene.
Specifically, the input of the task is a $T$-frame historical sequence of $N$ surround-view camera images $\{\mathcal{I}_{i,t} \in \mathbf{R}^{H_i \times W_i \times 3}\}$, where $i=1,...,N$ and $t=1,...,T$.
We also assume known sensor intrinsic parameters $\{K_{i}\}$ and extrinsic parameters $\{[R_i|t_i]\}$ in each frame.
The ground truth labels are the voxel states, including \textit{occupancy state} (``occupied'', ``free'', or ``unobserved'') and \textit{semantic label} (category, or ``unknow''). 
For example, a voxel on a vehicle is labeled as (``occupied'', ``vehicle''), and a voxel in the free space is labeled as (``free'', None).
Note that the 3D occupancy prediction framework also supports extra attributes as outputs, such as \textit{instance IDs} and \textit{motion vectors}; we leave them as future work.
\vspace{-4pt}

\subsection{Dataset Statistics}
We generate two 3D occupancy prediction datasets, Occ3D-nuScenes and Occ3D-Waymo. Occ3D-nuScenes contains 600 scenes for training, 150 scenes for validation, and 150 for testing, totaling 40,000 frames. It has 16 common classes with an additional general object (GO) class. Each sample covers a range of [-40m, -40m, -1m, 40m, 40m, 5.4m] with a voxel size of [0.4m,0.4m,0.4m].
Occ3D-Waymo contains 798 sequences for training, 202 sequences for validation, accumulating 200,000 frames. It has 14 known object classes with an additional GO class.
Each sample covers a range of [-80m, -80m, -1m, 80m, 80m, 5.4m], with an extremely fine voxel size of [0.05m, 0.05m, 0.05m].

Occ3D stands out when compared with other datasets, as shown in Table~\ref{table:datasets}. The indoor datasets NYUv2 and ScanNet lack surround images and consist of fewer sequences and frames. SemanticKITTI and KITTI-360, the two other outdoor datasets, also lack surround images, with the exception of KITTI-360's fisheye images. For the safety of autonomous driving, the general object class is particularly important, a feature that is not available in the SemanticKITTI and KITTI-360 datasets. Furthermore, Occ3D-Waymo is currently the 3D occupancy dataset with the most diverse scenarios, comprehensive labels, and the highest resolution among all open-source datasets.

\subsection{Dataset Construction Pipeline}

\begin{figure*}[t]
    \centering
    \includegraphics[width=0.99\linewidth]{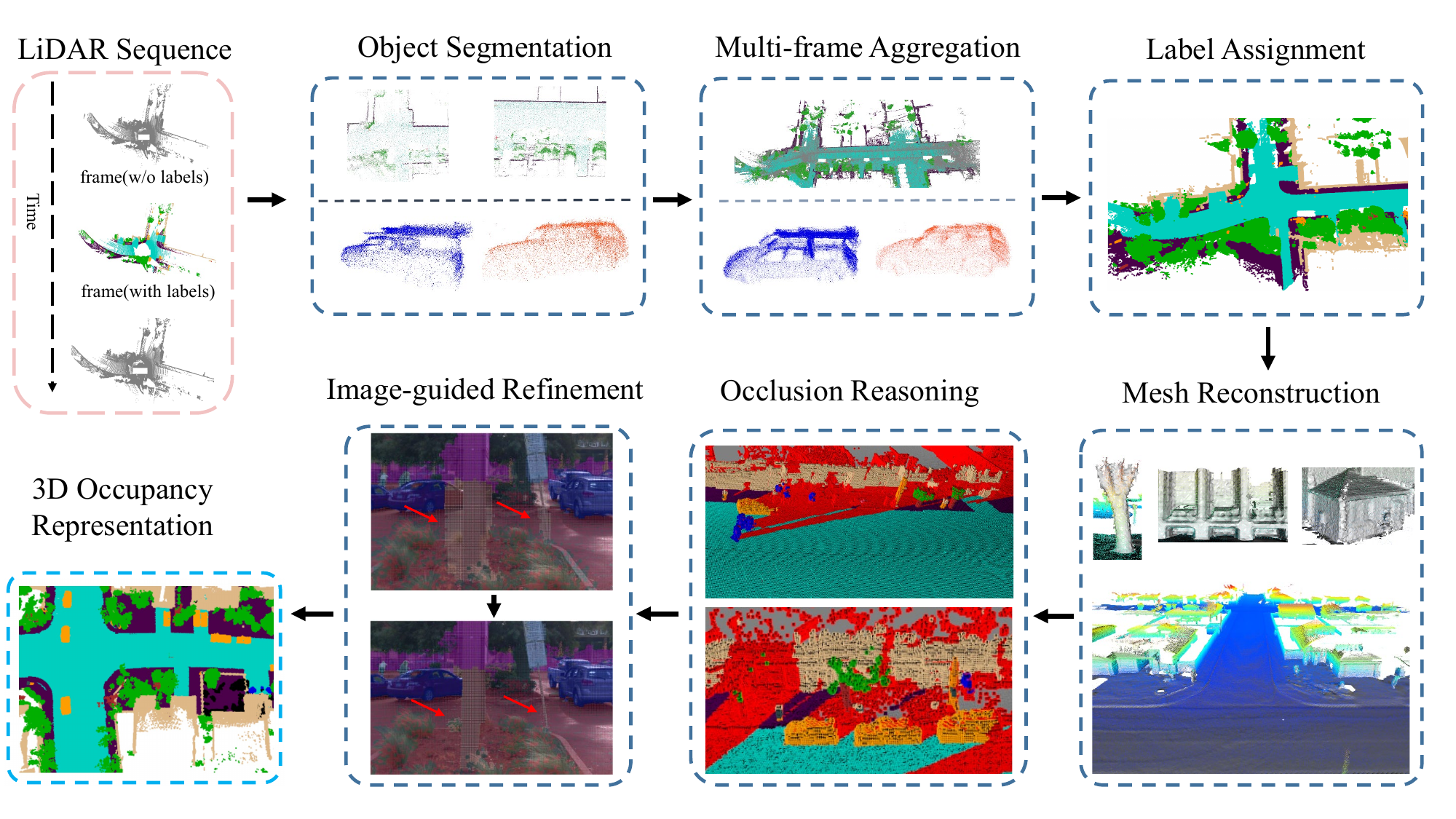}
    \caption{\textbf{Overview of the label generation pipeline.} The pipeline consists of three main steps: voxel densification, occlusion reasoning, and image-guided voxel refinement.Voxel densification consists of object segmentation, multi-frame aggregation, and label assignment. }
    \label{fig:pipeline}
\end{figure*}

Annotating 3D occupancy from images is impossible due to the lack of accurate depth and geometry. Therefore, we take advantage of LiDAR scans and their annotations to construct high-quality occupancy labels.
However, there are three primary hurdles: \textbf{sparsity}, \textbf{occlusion}, and \textbf{3D-2D misalignment}.
Sparsity refers to the fact that LiDAR scans are sparse, thereby hindering the acquisition of dense voxels.
Occlusion, on the other hand, is concerned with the identification of voxels that, once densified, become invisible in the current image view due to occlusion.
3D-2D misalignment pertains to the disparities when projecting the 3D voxels onto 2D images, often induced by sensor noises or pose errors.


Our proposed label generation pipeline addresses the above challenges, an overview is shown in Figure~\ref{fig:pipeline}. Initially, in \textbf{voxel densification}, we increase the density of the point clouds by performing multi-frame aggregation for both static and dynamic objects separately. Then we employ a K-nearest neighbor algorithm to assign labels to unlabeled points and utilize mesh reconstruction to perform hole-filling. 
Subsequently, we carry out \textbf{occlusion reasoning} from both LiDAR and camera perspectives, utilizing a ray-casting operation to label the occupancy state of each voxel.
Finally, misaligned voxels are eliminated through an \textbf{image-guided voxel refinement} process.
We provide pseudo-code and the hyper-parameters of each step in the Appendix.

\subsubsection{Voxel Densification}
LiDAR data is inherently sparse, to acquire dense point clouds: 1) We aggregate all points throughout the frames, treating dynamic objects and static background points separately; 2) We take advantage of unlabeled frames (which we'll refer to as non-keyframes) and use a K-Nearest Neighbors (KNN) algorithm to assign semantic labels; 3) In spite of frame aggregation, there persist holes on the object surfaces, we fill these holes with mesh reconstruction.

\noindent\textbf{Dynamic and static objects segmentation.}
Point clouds derived from individual frames are categorized into ``dynamic objects'' and ``static scenes''. 
The static scenes contain entities such as ground, buildings, and road signs that do not exhibit positional change over time. Dynamic objects, such as cars and pedestrians need to be segregated since naive temporal aggregation results in motion blur.


\noindent\textbf{Multi-frame aggregation.}
After segregating dynamic objects from static scenes, multi-frame aggregation is conducted separately on them.
For dynamic objects, we extract the points located within the annotated or tracked box and subsequently transform them from sensor coordinates to box coordinates. By concatenating these transformed points, we densify the point cloud of dynamic objects. 
For the static scene, we simply aggregate its points across time in the global coordinate system. 
The static scene is then fused with the aggregated dynamic objects in the current frame, thereby generating a single-frame dense point cloud.

\noindent\textbf{KNN for label assignment.}
The task of directly annotating each point in every frame is labor-intensive. Current datasets only annotate a selected portion of the frames - for instance, the Waymo dataset proceeds at a rate of 2Hz, whereas Lidar scans operate at a 10Hz frequency. 
To utilize the unlabeled frames, we employ the K-nearest neighbors (KNN) algorithm to assign semantic labels to each unlabeled point.
Specifically, for each point in the unlabeled frame, we find the K nearest keyframe points and assign the majority semantic label. 

\noindent\textbf{Mesh reconstruction.}
After multi-frame aggregation, the density of point clouds is still not enough to produce high-quality dense voxels: a smaller voxel size may lead to objects with many holes, while a larger voxel size could induce excessive smoothness. 
To mitigate these issues, we perform mesh reconstruction.
For non-ground categories, we optimize surfaces through VDBFusion~\cite{vizzo2022vdbfusion}, an approach for volumetric surface reconstruction based on truncated signed distance functions (TSDF). The flexibility and efficacy of VDBFusion surpass traditional methods such as Poisson surface reconstruction.
For the ground, VDBFusion fails as small ray angles result in incorrect TSDF values. We instead establish uniform virtual grid points and fit each local surface mesh using points within a small region. 
After reconstructing the meshes, dense point sampling is performed, and KNN is further adopted to assign semantic labels to the sampling points.

\subsubsection{Occlusion Reasoning for Visibility Mask}
We perform occlusion reasoning and introduce LiDAR visibility mask and camera visibility mask to further enhance our 3D occupancy prediction benchmark. 

\begin{figure}[t]
	\centering
	\subfloat[Occlusion Reasoning for Visibility Mask]{
		\label{fig:visibility}
		\centering
		\includegraphics[width=0.53\textwidth]{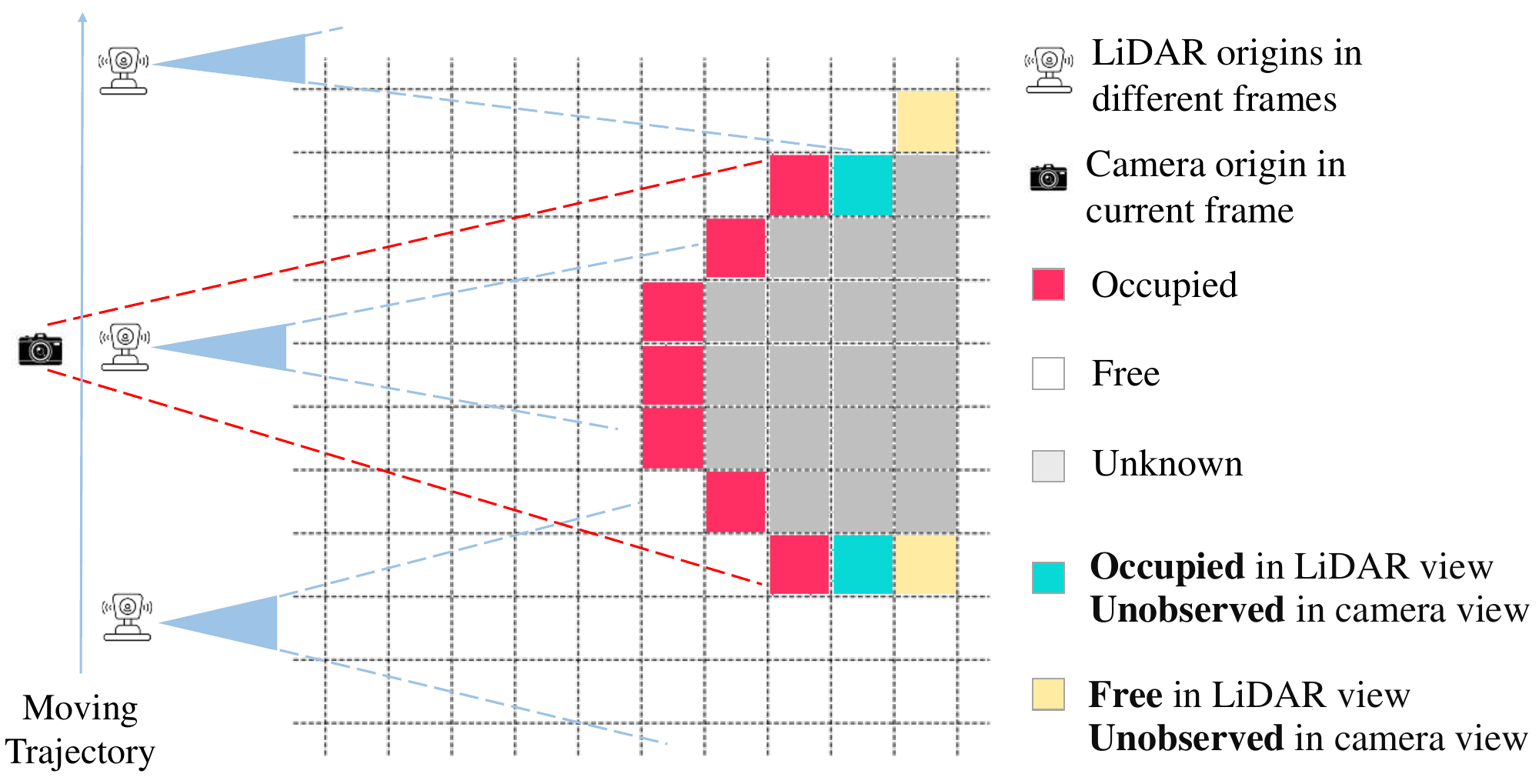}
	}
	\subfloat[Image-guided Voxel Refinement]{
		\label{fig:IGR}
		\centering
		\includegraphics[width=0.44\textwidth]{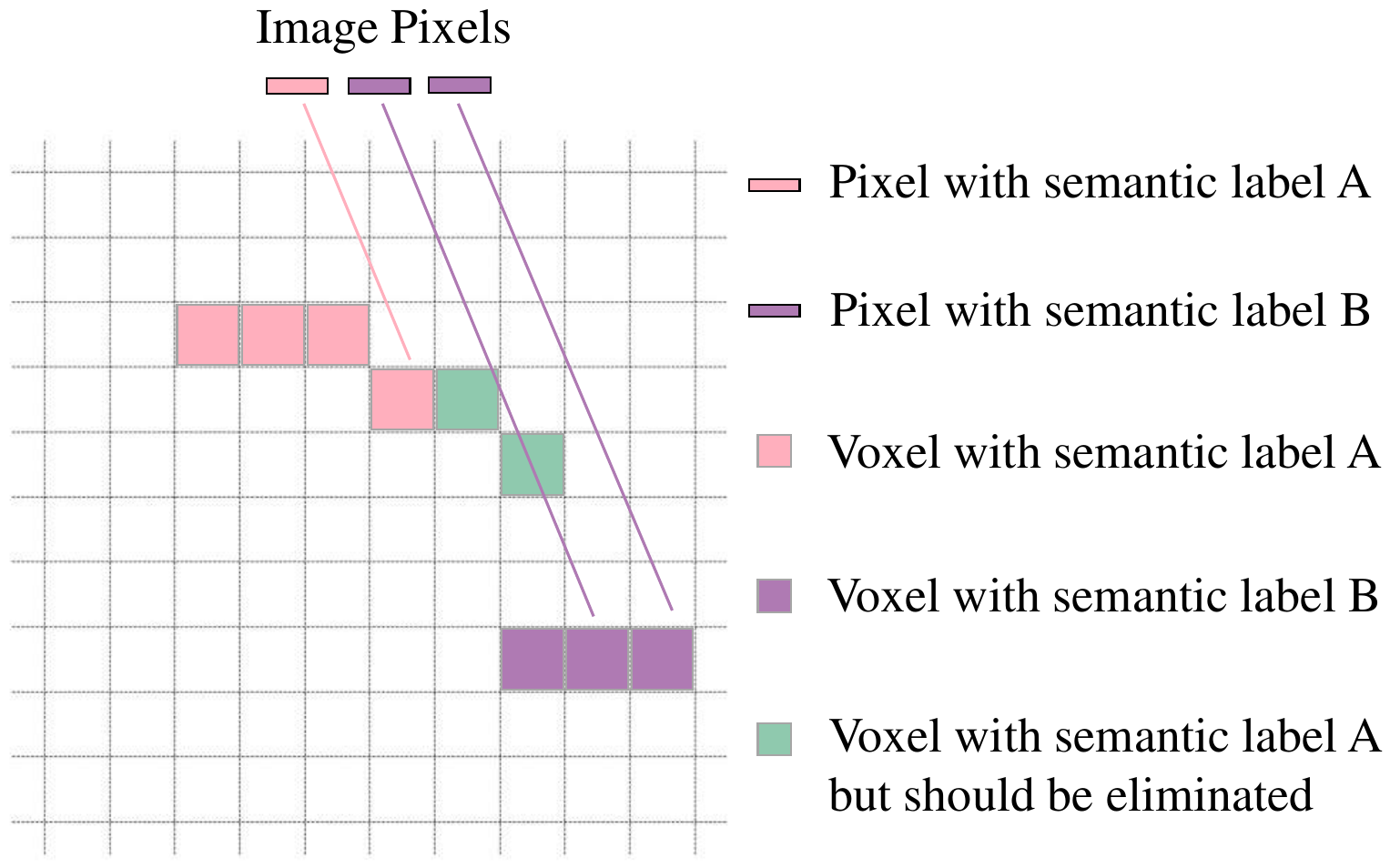}
	}
	\label{fig:ray}
	\caption{\textbf{Visibility and refinement.} (a) LiDAR visibility: a voxel is ``occupied'' if it reflects LiDAR (red voxels), or ``free'' if it is traversed through by a ray (white voxels); Camera visibility: Any voxel not scanned by camera rays is set to ``unobserved'' (blue and yellow voxels). (b) Image-guided voxel refinement: during ray casting, when the first voxel with the same semantic label as the pixel label is encountered, we set the previously traversed voxel states to ``free'' (green voxels).}
\end{figure}

\noindent\textbf{Aggregated LiDAR visibility mask.}
To obtain a 3D occupancy grid from aggregated LiDAR point clouds, a straightforward way is to set the voxels containing points to be ``occupied'' and the rest to ``free''. However, since LiDAR points are sparse, some occupied voxels are not scanned by LiDAR beams, and can be mislabeled as ``free''. To avoid this issue, we perform a ray casting operation to determine the visibility of each voxel, as shown in Figure~\ref{fig:visibility}. 
Concretely, we cast a ray from the sensor origins to each LiDAR point. A voxel is considered visible if it either reflects LiDAR points, or if it is traversed through by a ray. If neither condition is met, the voxel is classified as ``unobserved''.


\noindent\textbf{Camera visibility mask.} 
We connect each occupied voxel center with the camera origin, thereby forming a ray. Along each ray, we set the first occupied voxel as ``observed'', and the remaining as ``unobserved''.
Any voxel not scanned by camera rays is set to ``unobserved'' as well. Determining the visibility of a voxel is crucial for the evaluation of the 3D occupancy prediction task: evaluation is only performed on the ``observed'' voxels in both the LiDAR and camera views.

\subsubsection{Image-guided Voxel Refinement}
Influences such as LiDAR noise and pose drifts can cause the 3D shape of objects to appear larger than their actual physical dimensions. To rectify this, we further refine the dataset by eliminating incorrectly occupied voxels, guided by semantic segmentation masks of images. 
As shown in Figure \ref{fig:IGR}, to obtain the correspondence between 3D voxels and 2D pixels, we adopt a ray casting operation similar to the one in the previous section: connecting each occupied voxel center with the camera center to form a ray, and traverse the voxel that this ray passes through from near to far from the pixel origin. When the first voxel with the same semantic label as the pixel label is encountered, we set the previously traversed voxel states to ``free''. This step greatly improves the shape at object boundaries.





\vspace{2mm}
\section{Quality Check}

\begin{figure}[t]   
    \centering
    \subfloat[2D ROI]{
        \label{fig:quality1}
        \centering
        \includegraphics[width=0.32\textwidth]{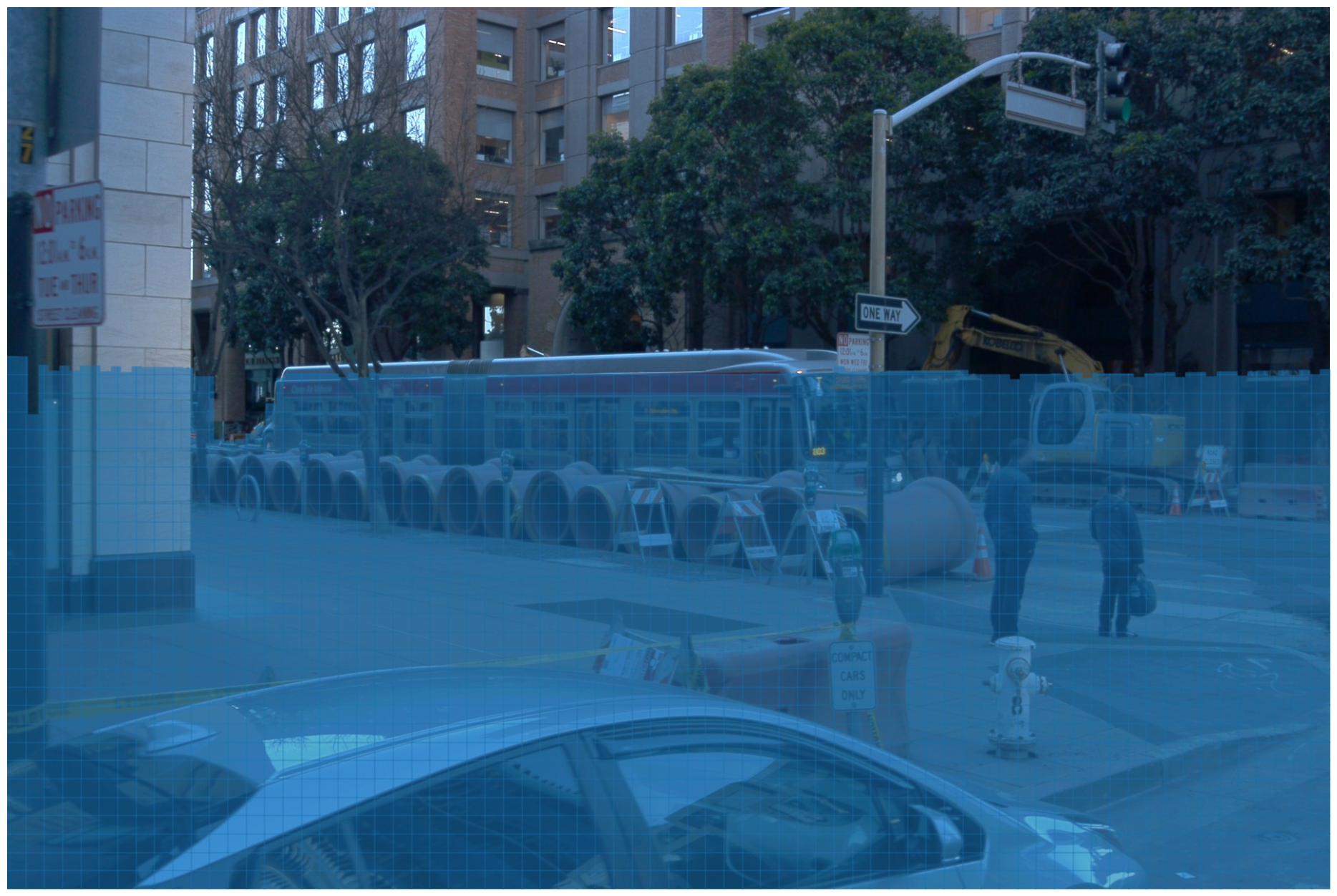}
    }
    \subfloat[2D pixel semantic label]{
        \label{fig:quality2}
        \centering
        \includegraphics[width=0.32\textwidth]{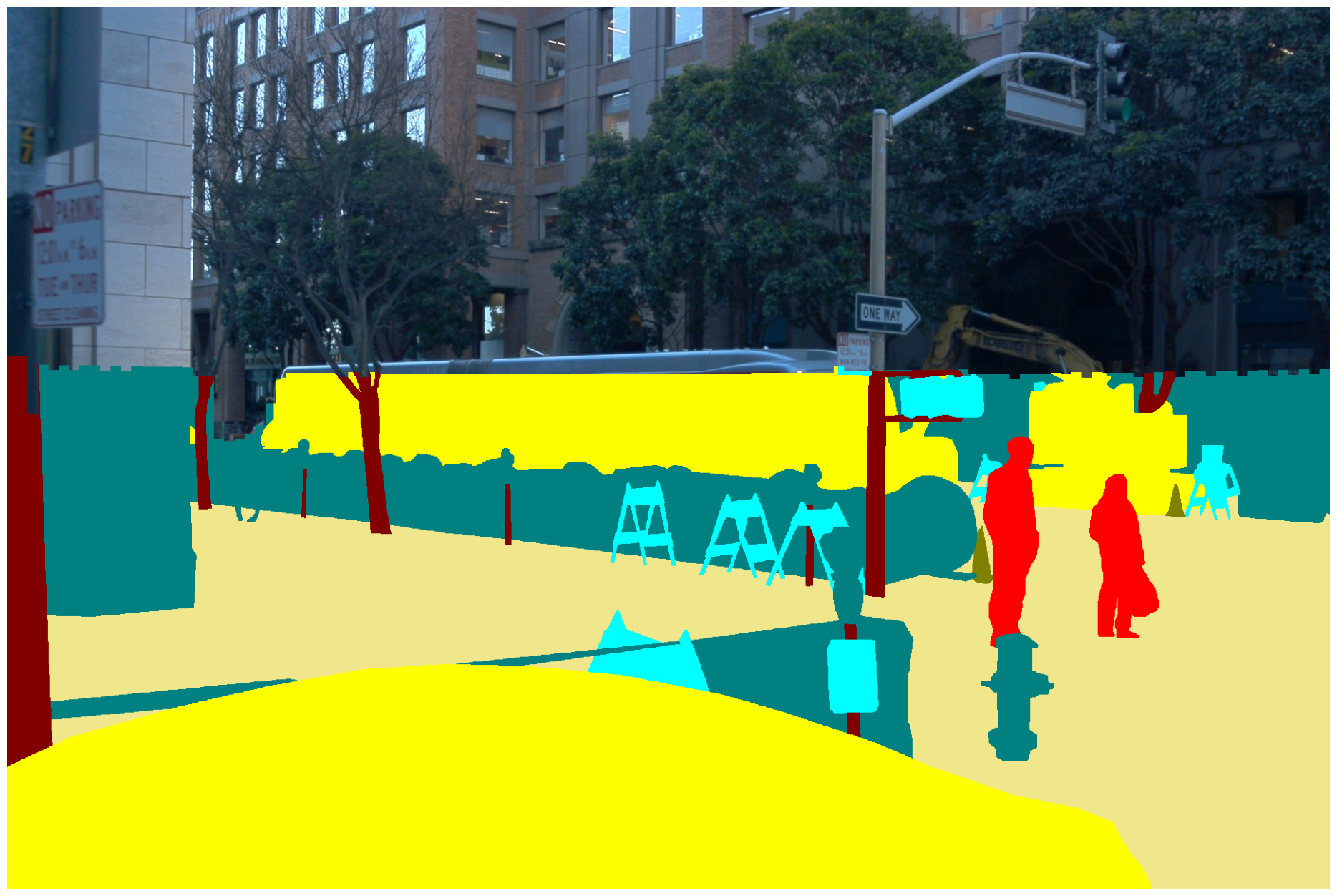}
    }
    \subfloat[3D voxel semantic label]{
        \label{fig:quality3}
        \centering
        \includegraphics[width=0.32\textwidth]{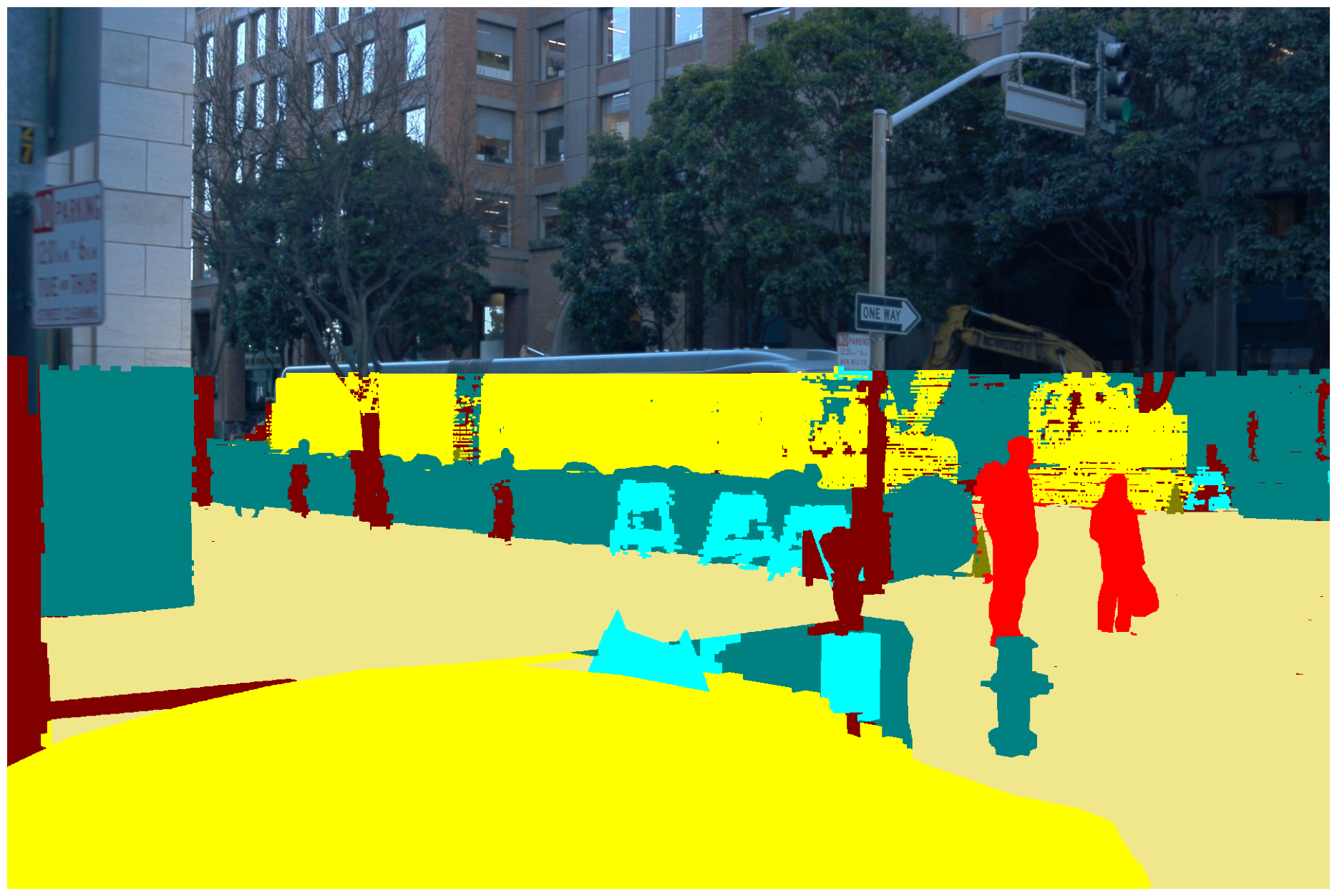}
    }
    \label{fig:quality_check}
    \caption{\textbf{3D-2D consistency} (a) 2D ROI within single-frame LiDAR scan range. (b) Semantic labels of a single image within the 2D ROI. (c) The reprojection of 3D voxel semantic labels onto the image within the 2D ROI.}
\end{figure}

Acquiring an occupancy representation that adheres to the complete shape of all objects is challenging. Therefore, evaluating the quality of the dataset and ensuring the effectiveness of each step in our pipeline is critical. To this end, we propose a method that evaluates the quality of occupancy by checking semantic consistency between 2D pixels and their corresponding voxel.
\vspace{2mm}
\subsection{3D-2D consistency}
Compared to 3D occupancy semantic labels obtained through aggregation and reconstruction, 2D semantic masks manually annotated by humans are highly accurate. Thus, we assess the quality of the dataset by verifying the 3D-2D consistency between semantic labels of 3D voxels and their corresponding 2D image pixels. We calculate 3D-2D consistently in three steps: filtering the 2D pixel region involved in consistency calculation for the current frame, identifying the corresponding 3D voxels of this pixel region, and finally, computing their 3D-2D semantic consistency. 

\noindent\textbf{2D ROI.}  2D images contain objects that are beyond the scanning range of the LiDAR sensor. 
When calculating 3D-2D consistency, we use the maximum range covered by a single LiDAR frame as the 2D Region of interest (ROI). Specifically, we project single-frame LiDAR points onto the 2D image coordinate system using LiDAR-to-camera transformation. Then, our algorithm traverses in the horizontal coordinate direction and selects the highest vertical coordinate of the projected points in each vertical column as the height of that column. As shown in Figure~\ref{fig:quality1}, all pixels below this height are treated as the 2D valid region involved in the consistency calculation. 

\noindent\textbf{3D label query.} After determining the 2D ROI in each image, we identify its corresponding 3D voxels for these regions. Since each voxel has a certain volume, directly projecting them onto a 2D image poses a multi-pixel association issue. Moreover, when the projection overlap occurs, determining the corresponding occlusion relationship becomes complicated. We instead query corresponding 3D voxels for each 2D image pixel.  Specifically, for each pixel in the selected region, we perform ray traversal and find the closest 3D voxel to the ray. 

\noindent\textbf{Metrics.} 
To evaluate the dataset quality, for each pixel in an image, we compare its semantic label with the semantic prediction of its corresponding 3D voxel. We adopt the standard Precision, Recall, Intersection-over-Union(IoU), and mean Intersection-over-Union(mIoU) metric.

\subsection{Quantitative Results}

Table \ref{table:dataset designs} shows the performance gain of each proposed step of our auto-labeling pipeline. The 3D-2D consistency is evaluated in a subset of Occ3D-Waymo.
Single-frame points (SFP) means that we only use a single-frame point cloud to calculate its 3D-2D consistency using the previously proposed method. As shown in the table, our method achieves high SFP precision and low recall. In addition to SFP, we aggregate points from multiple frames (MFP). Compared to SFP, MFP sees a significant improvement in recall, but its precision decreases to a certain extent, which is caused by the LiDAR noise and/or pose errors. 
Based on MFP, we study the effect of voxelization, which leads to better precision and recall. This further validates the effect of correction on pose inaccuracies. 
As mentioned before, a small voxel size results in objects containing many holes, while a larger voxel size leads to over smoothness. The former results in low recall, while the latter results in low precision. 
We use mesh reconstruction to alleviate the hole issue in objects caused by a small voxel size, which is reflected by the comparison between third row and fifth row in the table. 
Finally, we demonstrate that our proposed image-guided refinement indeed promotes the 3D-2D semantic consistency, shown in the last row.

\begin{table*}[t]
\scriptsize
\setlength{\tabcolsep}{2pt}
\newcommand{\classfreq}[1]{{~\tiny(\semkitfreq{#1}\%)}}  
\caption{\textbf{Quantitative results for design choices.} \textit{SFP}, single frame points; \textit{MFP}, aggregating points from unlabeled frames; \textit{VS}, short for voxel size; \textit{Mesh}, showcasing mesh reconstruction; and \textit{IGR}, denoting image-guided voxel refinement. The three numbers from top to bottom in each choice are IoU, recall, and precision for the specific class.} 
\def\mystrut{\rule{0pt}{1.5\normalbaselineskip}}
\centering
\begin{adjustbox}{width=0.99\columnwidth,center}
\begin{tabular}{l l l l l | c c c c c c c c c |c }
    \toprule
    SFP
    & MFP
    & VS
    & Mesh
    & IGR
    & vehicle
    & bicyclist
    & ped
    & sign
    & road
    & pole
    & cone
    & bicycle
    & building 
    & mIOU\\
    \midrule

        $\checkmark$ &  & - &   &  &\makecell{5.87 \\ 8.53 \\ 95.38} & \makecell{5.12 \\ 6.61 \\ 58.66} & \makecell{3.65 \\ 4.81 \\ 60.13} & \makecell{3.47 \\ 3.85 \\ 61.44}

         & \makecell{ 0.33 \\ 0.33 \\ 92.78} & \makecell{ 0.10 \\ 0.10 \\ 34.90} & \makecell{ 0.09 \\ 0.09 \\ 25.35} & \makecell{ 0.11 \\ 0.11 \\ 60.12} &  \makecell{ 0.34 \\ 0.34 \\ 66.93} & 13.32 \\

    \midrule
         $\checkmark$ & $\checkmark$ & - &  &  &\makecell{ 37.89 \\ 40.02 \\ 87.48} & \makecell{ 37.99 \\ 58.77 \\ 51.79} & \makecell{ 28.25 \\ 37.21 \\ 53.98} & \makecell{ 12.57 \\ 14.80 \\ 45.45} & \makecell{ 11.70 \\ 12.06 \\ 79.72} & \makecell{ 5.48 \\ 6.32 \\ 29.25 } & \makecell{ 3.51 \\ 3.99 \\ 22.37} & \makecell{ 6.01 \\ 6.45 \\ 46.76}  & \makecell{ 15.49 \\ 17.69 \\ 55.49} & 17.65 \\
        
    \midrule
         $\checkmark$ & $\checkmark$ & 0.1 &   &  &\makecell{ 75.23 \\ 91.20 \\ 81.12} & \makecell{ 38.66 \\ 87.00 \\ 41.03} & \makecell{ 30.78 \\ 60.90 \\ 38.37} & \makecell{ 33.77 \\ 56.80 \\ 45.45} & \makecell{ 56.30 \\ 67.35 \\ 95.53} & \makecell{ 30.58 \\ 55.97 \\ 40.27 } & \makecell{ 24.03 \\ 42.22 \\ 35.81} & \makecell{ 31.36 \\ 37.00 \\ 67.28}  & \makecell{ 49.85 \\ 68.66 \\ 64.53} & 41.17 \\

    \midrule
       $\checkmark$ &  $\checkmark$ & 0.1  & $\checkmark$ &   & \makecell{ 75.13 \\ 90.98 \\ 81.17} & \makecell{ 37.97 \\ 83.02 \\ 41.17} & \makecell{ 30.36 \\ 55.07 \\ 40.36} & \makecell{ 32.88 \\ 54.93 \\ 45.02} & \makecell{ 82.79 \\ 85.34 \\ 96.52} & \makecell{ 16.63 \\ 63.57 \\ 18.38} & \makecell{ 17.48 \\ 58.13 \\ 19.99} & \makecell{ 32.46 \\ 40.77 \\ 61.43}  & \makecell{ 52.27 \\ 77.81 \\ 61.43} & 41.99 \\
    \midrule
      $\checkmark$& $\checkmark$ & 0.05  &$\checkmark$ &   & \makecell{ 78.76 \\ 89.20 \\ 87.06} & \makecell{ 46.33 \\ 84.22 \\ 50.74} & \makecell{ 34.04 \\ 55.42 \\ 46.87} & \makecell{ 34.85 \\ 52.02 \\ 51.36} & \makecell{ 64.56 \\ 67.43 \\ 93.80} & \makecell{ 18.28 \\ 63.17 \\ 20.46} & \makecell{ 20.57 \\ 58.89 \\ 24.02} & \makecell{ 42.59 \\ 53.71 \\ 67.28} & \makecell{ 52.27 \\ 84.73 \\ 57.71} & 43.58\\
    \midrule
    $\checkmark$& $\checkmark$ & 0.05  & $\checkmark$ & $\checkmark$  & \makecell{ 88.82 \\ 91.34 \\ 96.98} & \makecell{ 76.89 \\ 92.12 \\ 82.31} & \makecell{ 47.54 \\ 63.78 \\ 65.11} & \makecell{ 50.18 \\ 61.20 \\ 73.59} & \makecell{ 71.97 \\ 75.11 \\ 94.51} & \makecell{ 31.77 \\ 73.80 \\ 35.80} & \makecell{ 32.09 \\ 62.44 \\ 39.76} & \makecell{ 66.40 \\ 77.67 \\ 82.07}  & \makecell{ 60.90 \\ 94.73 \\ 63.04} & 58.50 \\

    \bottomrule
\end{tabular}
\end{adjustbox}

\label{table:dataset designs}
\end{table*}

\section{Coarse-to-Fine Occupancy Network}
\begin{figure}[t]
    \centering
    \includegraphics[width=0.95\linewidth]{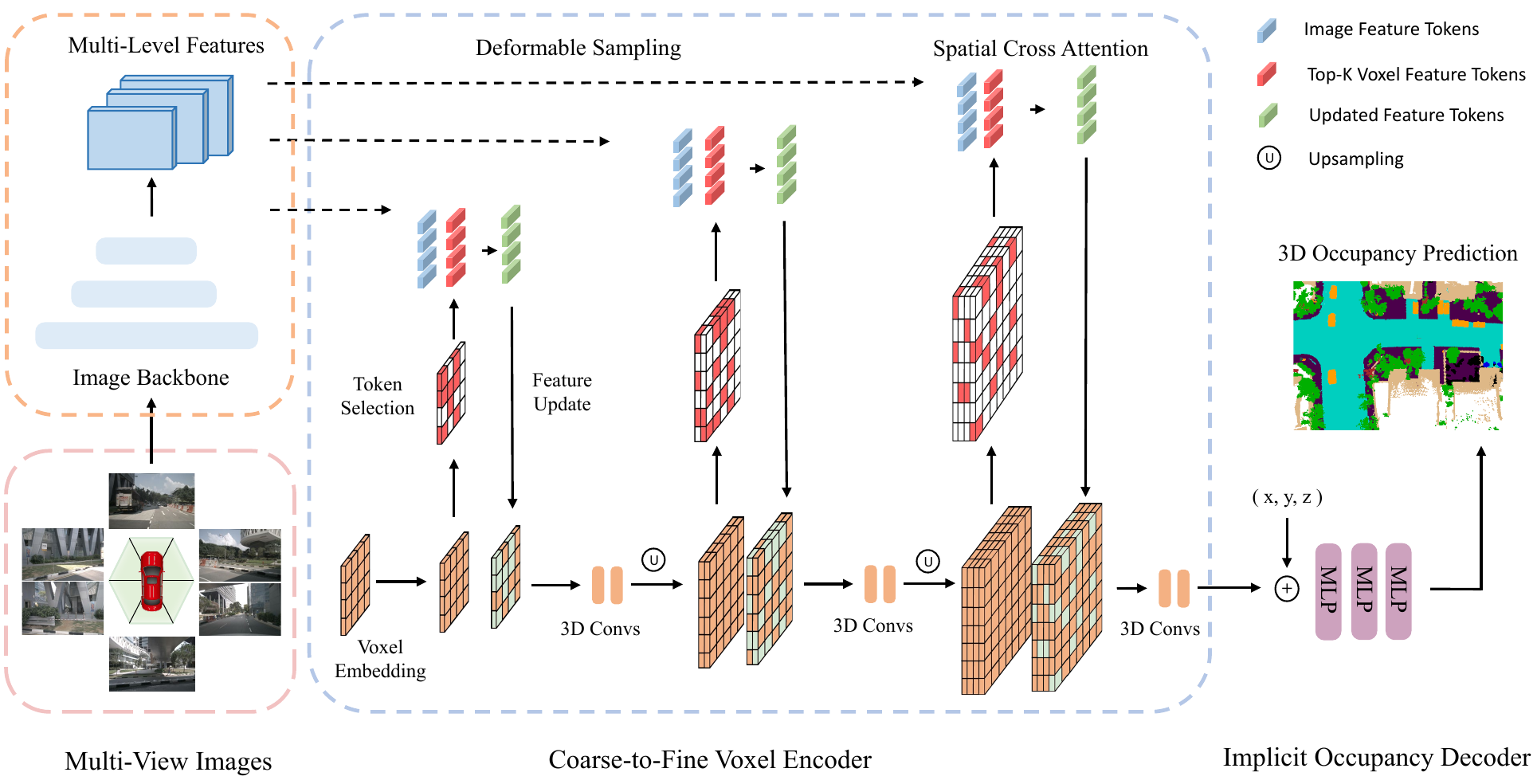}
    \caption{\textbf{The architecture of CTF-Occ network.} CTF-Occ consists of an image backbone, a coarse-to-fine voxel encoder, and an implicit occupancy decoder. }
    \label{fig:model}
\end{figure}

To deal with the challenging 3D occupancy prediction problem, we present a new transformer-based model named \textbf{C}oarse-\textbf{t}o-\textbf{F}ine \textbf{Occ}upancy (CTF-Occ) network.
An overview of CTF-Occ network is shown in Figure~\ref{fig:model}. First, 2D image features are extracted from multi-view images with an image backbone.
Then, 3D voxel queries aggregate 2D image features into 3D space via a cross-attention operation.
Our approach involves using a pyramid voxel encoder that progressively improves voxel feature representation through incremental token selection and spatial cross-attention in a coarse-to-fine fashion. This approach enhances the spatial resolution and refines the detailed geometry of objects, ultimately leading to more accurate 3D occupancy predictions.

\noindent \textbf{Incremental token selection.} 
The task of predicting 3D occupancy requires a detailed representation of geometry, but this can result in significant computational and memory costs if all 3D voxel tokens are used to interact with regions of interest in the multi-view images. Given that most 3D voxel grids in a scene are empty, we propose an incremental token selection strategy that selectively chooses foreground and uncertain voxel tokens in cross-attention computation. This strategy enables adaptive and efficient computation without sacrificing accuracy.
Specifically, at the beginning of each pyramid level, each voxel token is fed into a binary classifier to predict whether this voxel is empty or not. We use the binary ground-truth occupancy map as supervision to train the classifier. In our approach, we select the K-most uncertain voxel tokens for the subsequent feature refinement.

\noindent \textbf{Spatial cross attention.}
At every level of the pyramid, we first select the top-K voxel tokens and then aggregate the corresponding image features. In particular, we apply 3D spatial cross-attention \cite{li2022bevformer} to further refine the voxel features.

\noindent \textbf{Convolutional feature extractor.}
Once we apply deformable cross-attention to the relevant image features, we proceed to update the features of the foreground voxel tokens. Then, we use a series of stacked convolutions to enhance feature interaction throughout the entire 3D voxel feature maps. At the end of the current level, we upsample the 3D voxel features using trilinear interpolation.

\noindent \textbf{Occupancy decoder.}
The CTF voxel encoder generates voxelized feature output $V_{out}\in {\mathbb{R}^{W \times H \times L \times C}}$.
Then the voxel features $V_{out}$ are fed into several MLPs to obtain the final occupancy prediction $O\in {\mathbb{R}^{W \times H \times L \times C^{\prime}}}$, where $C^{\prime}$ is the number of the semantic classes. 
Furthermore, we introduce an implicit occupancy decoder that can offer arbitrary resolution output by utilizing implicit neural representations. The implicit decoder is implemented as an MLP that outputs a semantic label by taking two inputs: a voxel feature vector extracted by the voxel encoder and a 3D coordinate inside the voxel.

\section{Experiments}
To benchmark our proposed Occ3D datasets and our CTF-Occ model, we evaluate existing 3D occupancy prediction methods on Occ3D-nuScenes and Occ3D-Waymo.
\subsection{Experimental Setup}

\noindent \textbf{Dataset and Metrics.}\quad Occ3D-Waymo contains 1,000 publicly available sequences in total, where 798 scenes are for training and 202 scenes are for validation. The scene range is set from -40m to 40m along X and Y axis, and from -5m to 7.8m along Z axis. Occ3D-nuScenes contains 700 training scenes and 150 validation scenes. The occupancy scope is defined as -40m to 40m for X and Y axis, and -1m to 5.4m for the Z axis.
We choose a voxel size of $0.4$m to conduct our experiments on both two datasets. We adopt the metrics of Intersection-over-Union (IoU) and mean Intersection-over-Union(mIoU) to evaluate performance.

\noindent \textbf{Architecture.}\quad We extend two main-stream BEV models -- BEVDet~\cite{bevdet} and BEVFormer~\cite{li2022bevformer} to the 3D occupancy prediction task. We replace their original detection decoders with the occupancy decoder adopted in our CTF-Occ network and remain their BEV feature encoders. We employ ResNet-101~\cite{resnet} pretrained on FCOS3D~\cite{fcos3d} as the image backbone and the image size is resized to ($640 \times 960$) for Occ3D-Waymo and ($928 \times 1,600$) for Occ3D-nuScenes.  We also evaluate three existing 3D occupancy prediction methods -- MonoScene~\cite{cao2022monoscene}, TPVFormer~\cite{huang2023tri}, and OccFormer~\cite{zhang2023occformer} on our proposed Occ3D datasets. Additionally, we conduct experiments using LiDAR as an input on the Waymo dataset. ``LiDAR-Onl'' refers to adopting single frame LiDAR as input. Voxelization is applied with a voxel size of [0.1, 0.1, 0.4] on the x, y, and z axes respectively. Subsequently, a ResNet is employed to extract dense voxel features, which are then fed to the occupancy prediction head. The ``BEVFormer-Fusio'' method incorporates both camera and LiDAR inputs. We extract features from the same LiDAR branch and fuse them with the camera features captured by BEVFormer in the BEV space.

Our proposed CTF-Occ adopts a learnable voxel embedding with a shape of $200 \times 200 \times 256$. The voxel embedding will first pass through four encoder layers without token selection. There are three pyramid stage levels for the Occ3D-Waymo dataset, and the resolution of the z-axis in each stage is $8$, $16$, and $32$. The resolution of the z-axis in each stage for the Occ3D-nuScenes dataset is $8$ and $16$ for the two pyramid stages. Each stage contains one SCA layer and an incremental token selection module to choose K non-empty voxels with the highest scores. The top-k ratio for the incremental token selection strategy is set to 0.2 for all pyramid stages.

\noindent \textbf{Loss function.}\
To optimize the occupancy prediction, we use the OHEM loss for model training $\mathcal{L}_{occ} = \sum_{k}{W_k \mathcal{L}(g_k,p_k)}$,
where $W_k$, $g_k$, and $p_k$ represent the loss weight, the label, and the prediction result for the $k$-th semantic class.
In addition, we supervise the binary classification head in each pyramid level with binary voxel masks. The binary voxel masks are generated by processing the semantic occupancy label at each spatial resolution $s_i$ using $f(g,s_i)$, and the output of the binary classification head in the $i$-th level is denoted as $p_i$. The loss for the binary classification is defined as $\mathcal{L}_{bin} = \sum_{i}{\mathcal{L}(f(g,s_i),p_i)}$,
where $i$ represents the $i$-th pyramid level.

\begin{table*}[t]
\scriptsize
\caption{3D occupancy prediction performance on the Occ3D-nuScenes dataset. Cons. Veh represents construction vehicle and Dri. Sur is for driveable surface.} 
\setlength{\tabcolsep}{0.005\linewidth}
\newcommand{\classfreq}[1]{{~\tiny(\semkitfreq{#1}\%)}}  

\def\mystrut{\rule{0pt}{1.5\normalbaselineskip}}
\centering
\begin{adjustbox}{width=0.99\columnwidth,center}
\begin{tabular}{l | c c c c c c c c c c c c c c c c c |c}

    \toprule
    Method 
    & \rotatebox{90}{others} 
    & \rotatebox{90}{barrier}
    & \rotatebox{90}{bicycle} 
    & \rotatebox{90}{bus} 
    & \rotatebox{90}{car} 
    & \rotatebox{90}{Cons. Veh} 
    & \rotatebox{90}{motorcycle} 
    & \rotatebox{90}{pedestrian} 
    & \rotatebox{90}{traffic cone} 
    & \rotatebox{90}{trailer} 
    & \rotatebox{90}{truck} 
    & \rotatebox{90}{Dri. Sur} 
    & \rotatebox{90}{other flat} 
    & \rotatebox{90}{sidewalk} 
    & \rotatebox{90}{terrain} 
    & \rotatebox{90}{manmade} 
    & \rotatebox{90}{vegetation} 
    & \rotatebox{90}{mIoU}  \\
    \midrule
    
    MonoScene~\cite{cao2022monoscene}  & 1.75 & 7.23 & 4.26 & 4.93 & 9.38 & 5.67 & 3.98 & 3.01 & 5.90 & 4.45 & 7.17 & 14.91 & 6.32 & 7.92 & 7.43 & 1.01 & 7.65 & 6.06 \\
    TPVFormer~\cite{huang2023tri}  & 7.22 & 38.90 & 13.67 & 40.78 & 45.90 & 17.23 & 19.99 & 18.85 & 14.30 & 26.69 & 34.17 & 55.65 & 35.47 & 37.55 & 30.70 & 19.40 & 16.78 & 27.83 \\
    BEVDet ~\cite{bevdet}  & 4.39 & 30.31 & 0.23 & 32.26 & 34.47 & 12.97 & 10.34 & 10.36 & 6.26 & 8.93 & 23.65 & 52.27 & 24.61 & 26.06 & 22.31 & 15.04 & 15.10 & 19.38 \\
    OccFormer~\cite{zhang2023occformer}  & 5.94 & 30.29 & 12.32 & 34.40 & 39.17 & 14.44 & 16.45 & 17.22 & 9.27 & 13.90 & 26.36 & 50.99 & 30.96 & 34.66 & 22.73 & 6.76 & 6.97 & 21.93 \\
    BEVFormer~\cite{li2022bevformer}  & 5.85 & 37.83 & 17.87 & 40.44 & 42.43 & 7.36 & 23.88 & 21.81 & 20.98 & 22.38 & 30.70 & 55.35 & 28.36 & 36.0 & 28.06 & 20.04 & 17.69 & 26.88 \\
    \bf{CTF-Occ (Ours)}  & 8.09 & 39.33 & 20.56 & 38.29 & 42.24 & 16.93 & 24.52 & 22.72 & 21.05 & 22.98 & 31.11 & 53.33 & 33.84 & 37.98 & 33.23 & 20.79 & 18.0 & 28.53 \\

\bottomrule
\end{tabular}
\end{adjustbox}

\label{table:occ3d-nus}
\end{table*}
\vspace{6mm}

\begin{table*}[t]
\caption{3D occupancy prediction performance on the Occ3D-Waymo dataset. Cons. Cone represents the construction cone.} 
\setlength{\tabcolsep}{0.005\linewidth}
\newcommand{\classfreq}[1]{{~\tiny(\semkitfreq{#1}\%)}}  

\def\mystrut{\rule{0pt}{1.5\normalbaselineskip}}
\centering
\begin{adjustbox}{width=0.99\columnwidth,center}
\begin{tabular}{l| c  c c c c c c c c c c c c c c|c}
    \toprule
    Method 
    & \rotatebox{90}{GO} 
    & \rotatebox{90}{vehicle}
    & \rotatebox{90}{bicyclist} 
    & \rotatebox{90}{pedestrian} 
    & \rotatebox{90}{sign} 
    & \rotatebox{90}{traffic light} 
    & \rotatebox{90}{pole} 
    & \rotatebox{90}{Cons. Cone} 
    & \rotatebox{90}{bicycle} 
    & \rotatebox{90}{motorcycle} 
    & \rotatebox{90}{building} 
    & \rotatebox{90}{vegetation} 
    & \rotatebox{90}{tree trunk} 
    & \rotatebox{90}{road} 
    & \rotatebox{90}{sidewalk} 
    & \rotatebox{90}{mIoU}  \\
    \midrule
    BEVDet~\cite{bevdet}  & 0.13 & 13.06 & 2.17 & 10.15 & 7.80 & 5.85 & 4.62 & 0.94 & 1.49 & 0.0 & 7.27 & 10.06 & 2.35 & 48.15 & 34.12 & 9.88 \\
    TPVFormer~\cite{huang2023tri}   & 3.89 & 17.86 & 12.03 & 5.67 & 13.64 & 8.49 & 8.90 & 9.95 & 14.79 & 0.32 & 13.82 & 11.44 & 5.8 & 73.3 & 51.49 & 16.76\\
    BEVFormer~\cite{li2022bevformer}  & 3.48 & 17.18 & 13.87 & 5.9 & 13.84 & 2.7 & 9.82 & 12.2 & 13.99 & 0.0 & 13.38 & 11.66 & 6.73 & 74.97 & 51.61 & 16.76 \\
    \bf{CTF-Occ (Ours)}  & 6.26 & 28.09 & 14.66 & 8.22 & 15.44 & 10.53 & 11.78 & 13.62 & 16.45 & 0.65 & 18.63 & 17.3 & 8.29 & 67.99 & 42.98 & 18.73 \\
    \midrule

LiDAR-Only & 1.01 & 57.41 & 35.31 & 20.33 & 11.7 & 13.01 & 36.21 & 7.81 & 0.13 & 0.0 & 57.83 & 54.71 & 27.07 & 69.15 & 54.47 & 29.74 \\
BEVFormer-Fusion & 5.11 & 64.61 & 52.35 & 21.52 & 32.74 & 17.1 & 42.62 & 27.75 & 13.36 & 0.05 & 63.65 & 60.51 & 35.64 & 81.89 & 66.84 & 39.05 \\

\bottomrule
\end{tabular}
\end{adjustbox}
\label{table:occ3d-waymo}
\end{table*}

\subsection{Comparing with previous methods}
\noindent\textbf{Occ3D-nuScenes.}
Table~\ref{table:occ3d-nus} shows the performance of 3D occupancy prediction compared to related methods on the Occ3D-nuScenes dataset. It can be observed that our method performs better in all classes than previous baseline methods under the IoU metric. Our CTF-Occ surpass BEVFormer by $1.65$ mIoU. The observations are consistent with those in the Occ3D-Waymo dataset.    

\vspace{16pt}
\noindent\textbf{Occ3D-Waymo.} We compare the performance of our CTF-Occ network with state-of-the-art models on our newly proposed Occ3D-Waymo dataset. Results are shown in Table~\ref{table:occ3d-waymo}.
Our method outperforms previous  methods by remarkable margins, increasing the mIoU by $1.97$. Especially for some objects such as traffic cone and vehicle, our method surpasses the baseline method by $2.88$ and $10.23$ IoU respectively. This is because we capture the features in  the 3D voxel space without compressing the heigh, which will preserve the detailed geometry of objects. The results indicate the effectiveness of our coarse-to-fine voxel encoder.

\subsection{Ablation study}
\vspace{-5pt}
In this section, we ablate the choices of incremental token selection and OHEM loss. Table~\ref{tabel:ablation} shows the results. CC represents traffic cones and PED represents pedestrians. We focus on CC and PED to verify the effectiveness of our implementation on small objects. Both techniques improve performance. Using OHEM loss and top-k token selection produces the best performance. Without the OHEM loss, we only get $14.06$ mIoU. Combining the OHEM loss with a random token selection strategy achieves $16.62$ mIoU. Using an uncertain token selection strategy with OHEM loss achieve $17.37$ mIoU.
For token selection, uncertain selection and top-k selection are on par and they significantly outperform the random selection as expected. 

\begin{table*}[!ht]
    \centering
    \scriptsize
    \caption{Ablation study on our model components, performed on the Occ3D-Waymo dataset.\label{tabel:ablation}}
    \begin{adjustbox}{width=0.98\columnwidth,center}
        \begin{tblr}{Q[m,c,20mm] Q[m,c,20mm] Q[m,c,20mm] Q[m,c,20mm]  | Q[m,c,15mm] Q[m,c,15mm] | Q[m,c,15mm] }
            \hline[1pt]
             
           \SetCell[r=2]{m} OHEM Loss  & \SetCell[c=3]{m}  Token Selection Strategy & &   & \SetCell[c=2]{m} IoU &  & \SetCell[r=2]{m}  mIoU    \\
           \cline[0.5pt]{2-4}  \cline[0.5pt]{5-6}
            &random & uncertain & top-k & PED & CC &   \\
            \hline[1pt]
            & & & $\checkmark$ & $4.16$ & $10.03$ & $14.06$ \\
            $\checkmark$ & $\checkmark$ & & &  $5.07$ &  $12.95$ &  $16.62$ & \\
           $\checkmark$ &  & $\checkmark$ &  &  $6.27$ & $13.85$ & $17.37$  \\

           $\checkmark$ & & & $\checkmark$ & $7.04$ & $14.16$ & $18.43$ \\
            \hline[1pt]
        \end{tblr}
    \end{adjustbox}
    \vspace{-4pt}
    
    \label{tab:ablation_componet}
\end{table*}

\section{Conclusion}
\label{sec:conclusion}
We present Occ3D, a large-scale high-quality 3D occupancy prediction benchmark for visual perception. Meanwhile, we present a rigorous label generation protocol and a new model CTF-Occ network for the 3D occupancy prediction task. They are publicly released to facilitate future research. 
\vspace{16pt}


\noindent\textbf{Limitations.}
Although we meticulously design the dataset generation pipeline to significantly enhance its quality, there are several ways to achieve further improvement:
\begin{enumerate}[i.]
    \item Sensor Calibration Error: Since we use LiDAR scans to construct high-quality occupancy labels for camera perception, the calibration between LiDAR and cameras becomes critical. Conducting multi-frame aggregation also relies on precise sensor calibration.
    
    \item Dynamic and Deformable Objects: For dynamic objects, we extract the points located within the box and aggregate them. However, some dynamic objects may not have box annotations, such as running animals, and some objects may not satisfy the rigid body assumption, like a person swinging their arms. There will be motion blur problems in these cases.
    
    \item General Objects: Both the nuScenes and Waymo datasets only annotate limited categories. Out-of-vocabulary objects such as trash cans and traffic cones are all regarded as general objects. Further human annotation to provide fine-grained details will help in reproducing an intelligence with unbounded understanding and benefit auto-driving research.
\end{enumerate}

\noindent\textbf{Acknowledgments.}
This work is supported by the National Key R\&D Program of China (2022ZD0161700).

\small{
\bibliographystyle{plainnat}
\bibliography{egbib}
}
\clearpage
\appendix

\section*{\huge{Appendix}}
\vspace{8pt}
\section{Occ3D Dataset}
We publish the Occ3D dataset, benchmark, develop kit, data format
and annotation instructions at our website \href{https://tsinghua-mars-lab.github.io/Occ3D/}{Page-Occ3D}. It is our priority to protect
the privacy of third parties. We bear all responsibility in case of
violation of rights, etc., and confirmation of the data license. 



\textbf{Terms of use, privacy and License.} The Occ3D-nuScenes and Occ3D-Waymo dataset is published under \href{https://en.wikipedia.org/wiki/MIT_License}{MIT} license, which means everyone can use this dataset for non-commercial research purpose. The original nuScenes dataset is released under the \href{https://creativecommons.org/licenses/by-nc-sa/4.0/legalcode}{CC BY-NC-SA 4.0}. The original Waymo dataset is released under the \href{https://waymo.com/open/terms/}{Waymo Dataset License Agreement for Non-Commercial Use (August 2019)} License. 

\textbf{Data maintenance.} Data is stored in Google Drive for global users, and the Occ3D-nuScenes is stored in 
\href{https://drive.google.com/drive/folders/1wZ-8OI1IJkrXo6BudFSGmaKXBUYQ3ts_}{here} and the link
for Occ3D-Waymo is stored in 
\href{https://drive.google.com/drive/folders/13WxRl9Zb_AshEwvD96Uwz8cHjRNrtfQk}
{here}. We will 
maintain the data for a long time and check the data accessibility on a regular basis.

\textbf{Benchmark and code.} \href{https://eval.ai/web/challenges/challenge-page/2045/leaderboard/4838}{Benchmark-Occ3D-nuScenes} provides benchmark results of Occ3D-nuScenes. The label generation code will be released upon acceptance. 


\textbf{Data statistics.} For Occ3D-Waymo, there are 798 scenes for training, 202 scenes for valuation, 150 scenes for testing, and 200,000 frames in total. For Occ3D-nuScenes, there are 700 scenes for training, 150 scenes for valuation, 150 scenes for testing, and 40,000 frames in total.


\textbf{Limitations.} The proposed label generation pipeline does not achieve perfect reconstruction and is limited in several ways: it relies on precise sensor calibration, it does not handle deformable objects, etc. Future work will aim to address these issues.


\section{Mesh Reconstruction}
\begin{figure}[!hb]
    \centering
    \subfloat[]{
    \includegraphics[width=0.48\textwidth]{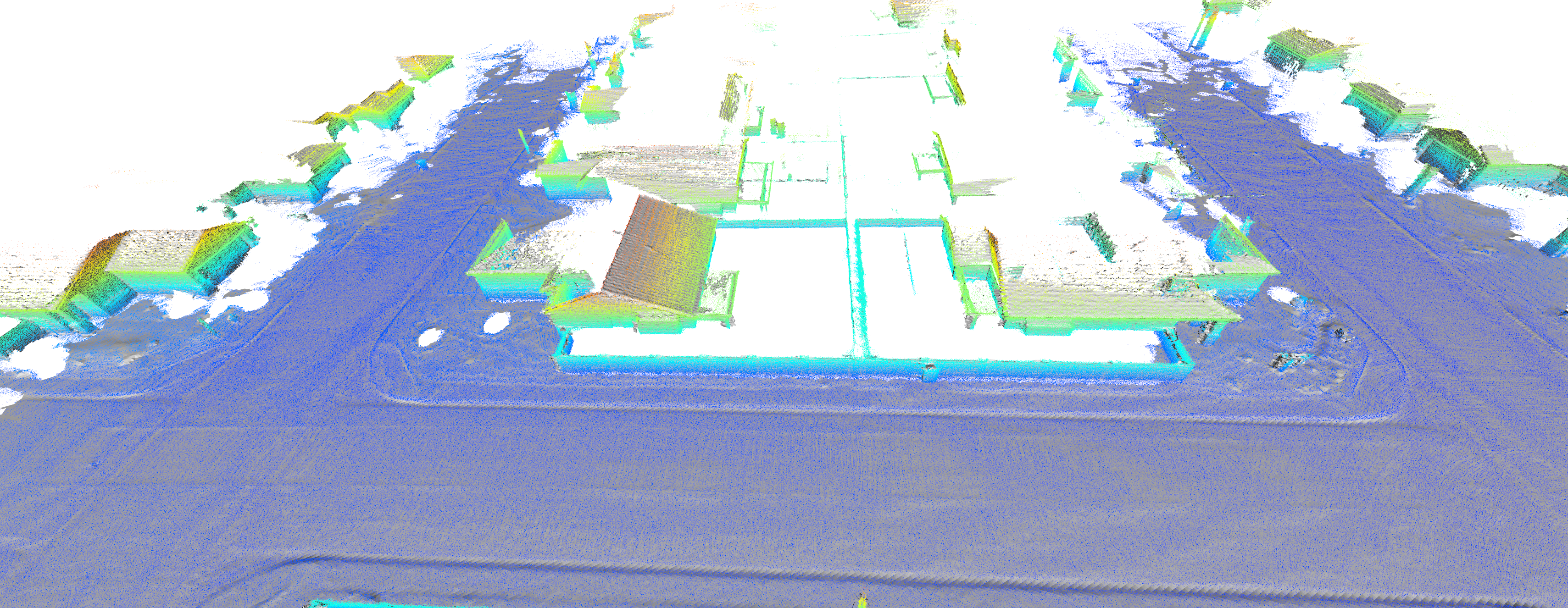}
    \label{fig:mesh_a}
    }
    \subfloat[]{
    \includegraphics[width=0.48\textwidth]{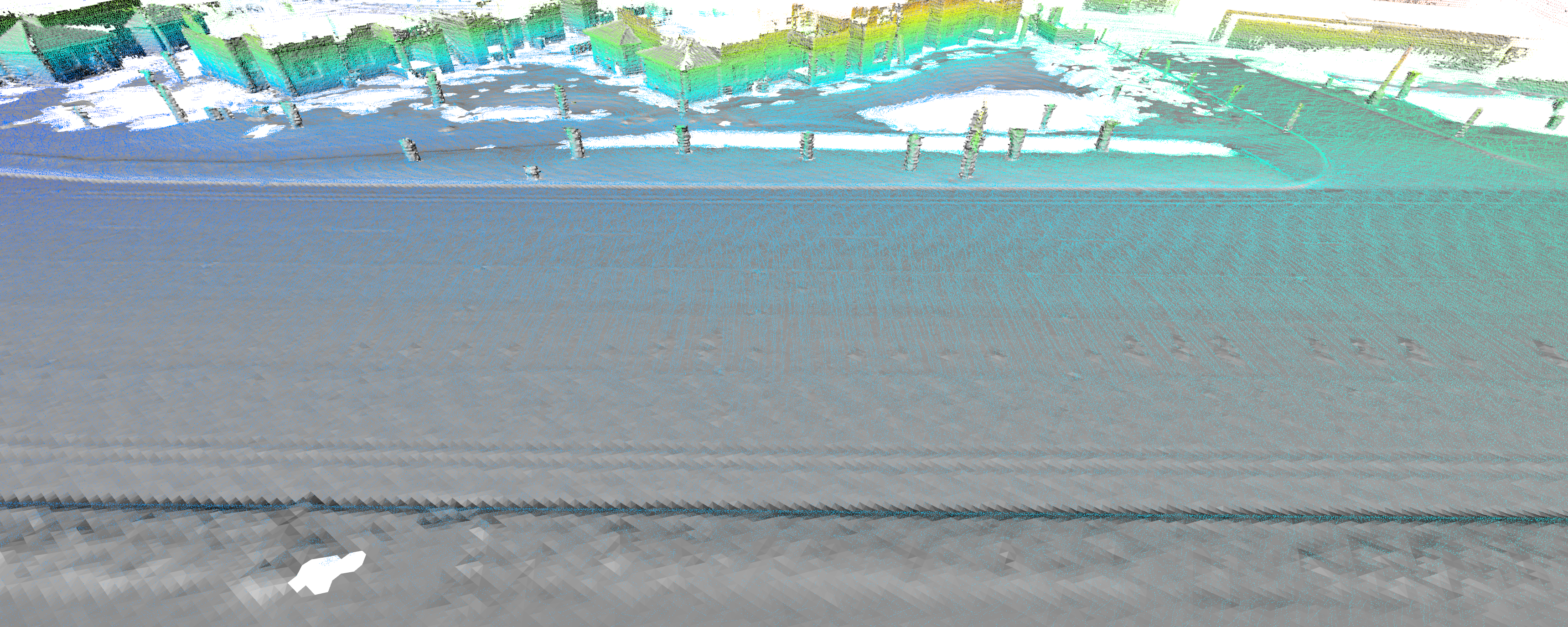}
    \label{fig:mesh_b}
    }
    \quad
    \subfloat[]{
    \includegraphics[width=0.96\textwidth]{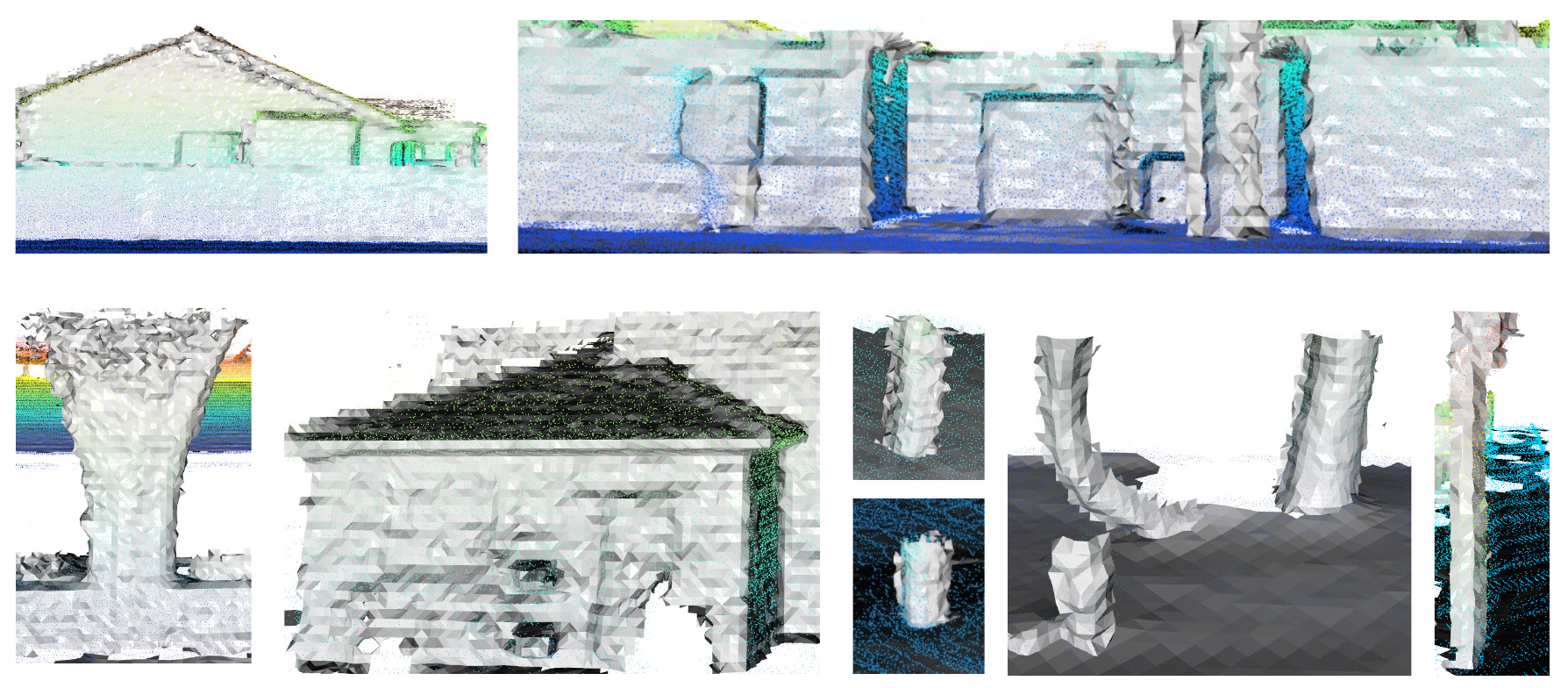}
    \label{fig:mesh_c}
    }
    \caption{\textbf{Mesh reconstruction visualization.} (a) and (b): A couple of scenes after mesh reconstruction. Blue points are the aggregated points, and the gray surface is the reconstructed mesh. (c): Some reconstructed objects, including houses, walls, trees, fire hydrants, and poles.}
    \label{fig:vis}
\end{figure}

We apply mesh reconstruction on the aggregated point cloud, and then resample to create a denser voxel representation. In Figures \ref{fig:mesh_a} and \ref{fig:mesh_b}, the color points represent the aggregated point cloud. It is evident that there are still holes between the points in the original point cloud, which, if converted directly to voxels, would result in many holes with a small voxelization size. After the mesh reconstruction, not only are these holes eliminated, but noisy areas are also effectively smoothed out.
Figure \ref{fig:mesh_c} shows the results of mesh reconstruction on some objects, including houses, walls, trees, fire hydrants, and poles. As can be observed from the figure, mesh reconstruction is able to effectively perform high-quality surface reconstruction on areas of the objects where point clouds are present.

\vspace{4mm}
\section{General Objects}
\begin{figure}[!h]
    \centering
    \captionsetup{type=figure}
    \includegraphics[width=0.96\textwidth]{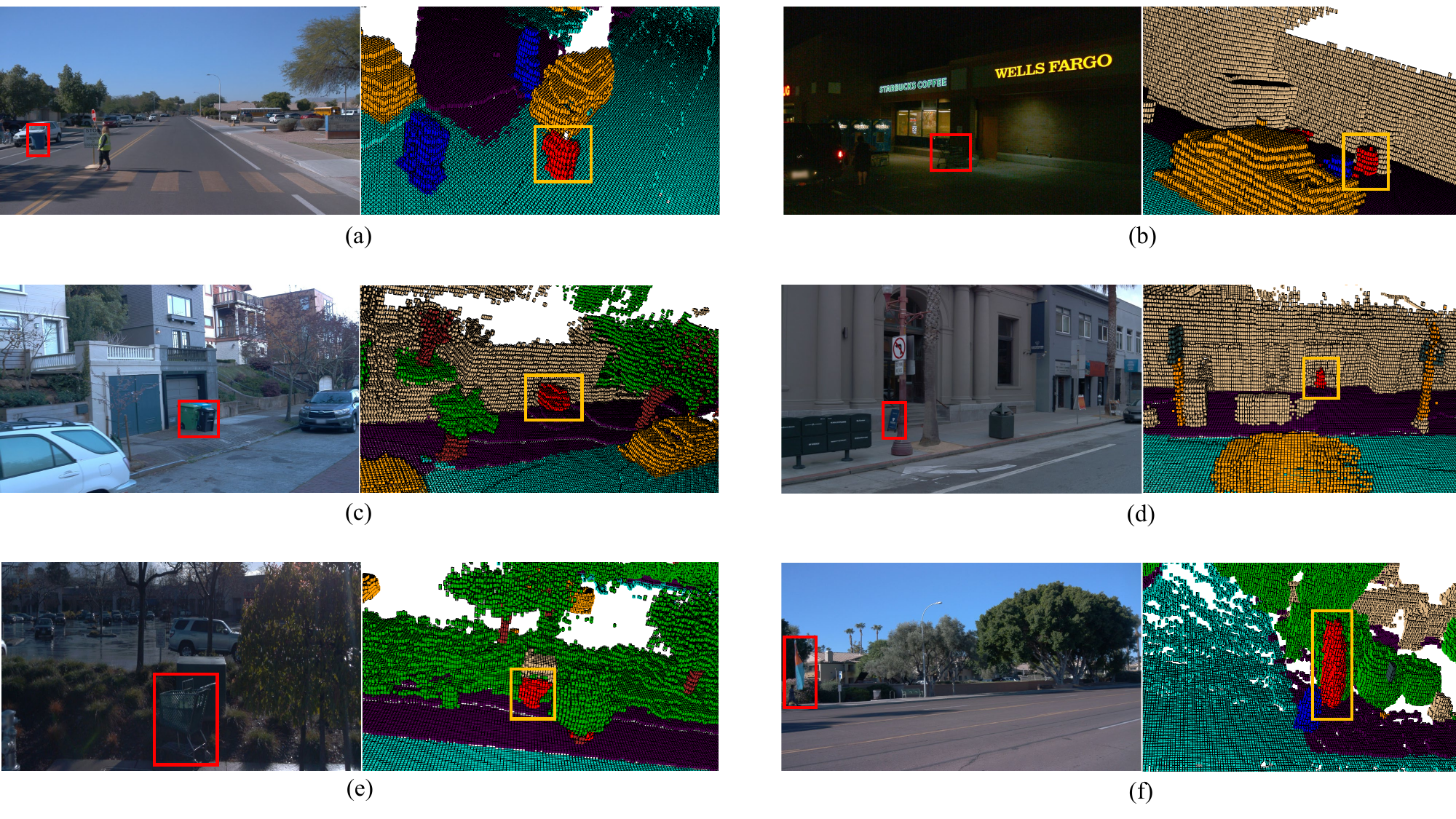}
    \caption{\textbf{General objects in our Occ3D benchmark.} We mark the general objects with red boxes in the camera view and yellow boxes in the voxel view.} 
    \label{fig:go}
\end{figure}

One of the key advantages of the 3D semantic occupancy prediction task is the potential to handle General Objects (\textbf{GOs}), or unknown objects. Different from 3D object detection which pre-defines categories of all the objects, 3D occupancy prediction handles arbitrary objects with occupancy grids and semantics. The geometries of objects are generally represented by voxels including out-of-vocabulary objects labeled as (``occupied", ``unknown"). This ability to represent and detect general objects makes the task more general and suitable for autonomous driving perception. Thus, we present a method using the clustering algorithm to handle ``unknown" objects.

We showcase several examples of GOs in our Occ3D benchmark in Figure~\ref{fig:go}. Figure~\ref{fig:go}(a) and (c) depict a dustbin, while Figure~\ref{fig:go}(b) and (e) show a shopping cart. Figure~\ref{fig:go}(d) displays a board on the sidewalk. Figure~\ref{fig:go}(f) features a flying banner. In each case, the voxels within the bounding box represent the corresponding GO.

\vspace{4mm}
\section{Visibility}
\textbf{Ray casting.} Both the Aggregated LiDAR and Camera visibility calculation heavily depend on a ray-casting algorithm, which is described in detail in Algorithm~\ref{alg:ray_casting}. The algorithm's execution is divided into two stages: the initialization phase (Lines 4 to 32) and the incremental traversal phase (Lines 33 to 65).

During the initialization phase, several parameters are determined: the ray direction $step$, the starting voxel coordinates $cur\_voxel$, the ending voxel coordinates $last\_voxel$, the first voxel boundary $tMax$, and $tDelta$ which defines the distance traversed along the ray when crossing a voxel.
The algorithm initiates at the ray's origin. It traverses each voxel in an interval order and continues looping until it encounters the last voxel within the specified range.

The $EPS$ hyper-parameter, set to $1e-9$, is used to nudge the start and end points of the ray slightly inside the traversed voxels to handle edge cases where a ray exactly intersects a voxel boundary. The $DISTANCE$ hyper-parameter, set to 0.5, determines the traversal threshold for the voxel grid, ensuring the ray stops casting when it exceeds the grid.



\begin{algorithm}[H]
\tiny 
\SetAlgoLined
\KwData{ray\_start $\in$ List[3], ray\_end $\in$ List[3], pc\_range $\in$ List[6], voxel\_size $\in$ List[3],  spatial\_shape $\in$ List[3]}
\KwResult{cur\_voxel $\in$ List[3]}
\SetKwProg{Fn}{Function}{:}{}
\Fn{ray\_casting}{
    $new\_ray\_start[0:3] \gets ray\_start[0:3]  - pc\_range[0:3]$\\
    $new\_ray\_end[0:3] \gets ray\_end[0:3]  - pc\_range[0:3]$\\
    \For{k in 0 to 2}{
        $ray[k] \gets new\_ray\_end[k] - new\_ray\_start[k]$\\
        \uIf{$ray[k] \geq0$}{
            $step[k] \gets 1$
        }
        \uElse{
            $step[k] \gets -1$
        }
        \eIf{$ray[k] \neq0$}{
            $tDelta[k] \gets (step[k] * voxel\_size[k]) / ray[k]$
        }{
            $tDelta[k] \gets FLOAT\_MAX$
        }
        $new\_ray\_start[k] \gets new\_ray\_start[k] + step[k]*voxel\_size[k]*EPS$\\
        $new\_ray\_end[k] \gets new\_ray\_end[k] - step[k]*voxel\_size[k]*EPS$\\
        $cur\_voxel [k] \gets \lfloor new\_ray\_start[k] / voxel\_size[k] \rfloor$\\
        $last\_voxel[k] \gets \lfloor new\_ray\_end[k] / voxel\_size[k] \rfloor$\\
    }

    \For{k in 0 to 2}{
        \eIf{$ray[k] \neq0$}{
            $cur\_coordinate \gets cur\_voxel [k]*voxel\_size[k]$\\
            \eIf{$step[k] < 0$ \textbf{and} $cur\_coordinate < new\_ray\_start[k]$}{
                $tMax[k] \gets cur\_coordinate\;$
            }{
                $tMax[k] \gets cur\_coordinate + step[k]*voxel\_size[k]\;$
            }
            $tMax[k] \gets (tMax[k] - new\_ray\_start[k]) / ray[k]\;$
        }{
            $tMax[k] \gets FLOAT\_MAX\;$
        }
    }    
    \While{$step * (cur\_voxel - last\_voxel)<DISTANCE$
   }{
        /* Determine the axis to move based on tMax comparison */\\
        \eIf{tMax[0] < tMax[1]}{
            \eIf{tMax[0] < tMax[2]}{
                $cur\_voxel [0] \gets cur\_voxel [0] + step[0]\;$
                
                \If{cur\_voxel [0] < 0 or cur\_voxel [0] $\geq$ spatial\_shape[0]}{\textbf{break}}
                $tMax[0] \gets tMax[0] + tDelta[0]\;$
            }{
                $cur\_voxel [2] \gets cur\_voxel [2] + step[2]\;$
                
                \If{cur\_voxel [2] < 0 or cur\_voxel [2] $\geq$ spatial\_shape[2]}{\textbf{break}}
                $tMax[2] \gets tMax[2] + tDelta[2]\;$
            }
        }{
            \eIf{tMax[1] < tMax[2]}{
                $cur\_voxel [1] \gets cur\_voxel [1] + step[1]\;$
                
                \If{cur\_voxel [1] < 0 or cur\_voxel [1] $\geq$ spatial\_shape[1]}{\textbf{break}}
                
                $tMax[1] \gets tMax[1] + tDelta[1]\;$
            }{
                $cur\_voxel [2] \gets cur\_voxel [2] + step[2]\;$
                
                \If{cur\_voxel [2] < 0 or cur\_voxel [2] $\geq$ spatial\_shape[2]}{\textbf{break}}
                
               $tMax[2] \gets tMax[2] + tDelta[2]\;$
            }
        }
    yield $cur\_voxel \;$
    }
}
\caption{Ray Casting}
\label{alg:ray_casting}
\end{algorithm}

\textbf{Aggregated LiDAR visibility.} The calculation of aggregated LiDAR visibility is described in Algorithm~\ref{alg:LiDAR_visibility}. The term $points$ denotes the aggregated point cloud, and $point origin$ stands for the corresponding LiDAR origin. Initially (Line 2), $voxel\_state$ is set to $NOT\_OBSERVED$, and $voxel\_label$ is initialized as $FREE\_LABEL$.
In Lines 12-13, for each voxel related to a point, the voxel occupancy counts $voxel\_occ\_count$ is accumulated by one, and the $voxel\_label$ is assigned the label of the current point. For any voxel that the ray passes through, the voxel free count $voxel\_free\_count$ is accumulated.
Finally, the state of voxels with $voxel\_free\_count$ greater than zero is set as $FREE$, and those with $voxel\_occ\_count$ greater than zero are set as $OCCUPIED$.
Despite a large number of points, often up to 2 million, the computation time is optimized to around 10 milliseconds by utilizing parallel processing on GPU, as shown in Line 5.

\begin{algorithm}[H]
\tiny 
\SetAlgoLined
\KwData{$points\_origin \in Tensor(N, 3), points \in Tensor(N, 3), points\_label \in Tensor(N,), pc\_range \in List[6], voxel\_size \in List[3],  spatial\_shape \in List[3]$}
\KwResult{$voxel\_state \in Tensor(H,W,Z), voxel\_label \in Tensor(H,W,Z)$}
\SetKwProg{Fn}{Function}{:}{}
\Fn{$calculate\_LiDAR\_visibility$}{
    Initialize $voxel\_occ\_count \in Tensor(H,W,Z), voxel\_free\_count \in Tensor(H,W,Z)$\\
    $voxel\_state \gets NOT\_OBSERVED, voxel\_label \gets FREE\_LABEL, voxel\_occ\_count \gets 0, voxel\_free\_count \gets 0 $\\
    Filter points, $points\_origin$, and $points\_label$ within $pc\_range $\\
    \For{i in 0 to N}{
        $ray\_start \gets points[i]$ \\
        $ray\_end \gets points\_origin[i]$\\
        
        \For{k in 0 to 2}{
            $ target\_voxel[k] \gets  \lfloor\frac{(ray\_start[k]-pc\_range[k])}{voxel\_size[k]}\rfloor$ \\
        } 

        \uIf{target\_voxel  $\in$ spatial\_shape}{
            $atomicAdd(voxel\_occ\_count[target\_voxel], 1)$\\
            $voxel\_label[target\_voxel] \gets points\_label[i]$\\
        }

        \For{voxel\_index in ray\_casting(ray\_start, ray\_end, pc\_range, voxel\_size, spatial\_shape)}{            
            $atomicAdd(voxel\_free\_count[voxel\_index], 1)$\\
        } 
    }
    $voxel\_state[voxel\_free\_count $\textgreater$ 0] \gets FREE$\\
    $voxel\_state[voxel\_occ\_count $\textgreater$ 0] \gets OCCUPIED$\\
}
\caption{Aggregated LiDAR Visibility}
\label{alg:LiDAR_visibility}
\end{algorithm}

\textbf{Camera visibility.} The calculation of camera visibility is described in Algorithm~\ref{alg:Camera_visibility}.
In Line 3, $update\_voxel\_state$ is initialized to $NOT\_OBSERVED$, and then some voxels marked as $OCCUPIED$ and $FREE$ under the LiDAR view are further assigned to $NOT\_OBSERVED$.
For each pixel in each camera image, a virtual point is generated at a significant distance away, as illustrated in Lines 5-9. Then points are transformed from the image coordinates to the ego coordinate system (Line 10). The camera origin serves as the origin for the virtual point, and is similarly transformed to the ego coordinate system in Lines 12-15. In Lines 24-34, the $update\_voxel\_state$ for voxels traversed by the pixel ray is assigned the same value as $voxel\_state$. 
The $DEPTH\_MAX$ hyper-parameter, which is set to $1e3$, acts as a surrogate for substantial depth. To enhance computational efficiency, each ray's operation executes concurrently on a GPU, as demonstrated in Line 21.

\begin{algorithm}[H]
\tiny 
\SetAlgoLined
\KwData{$Image$ $\in$ Tensor(K, h, w), $P_{cam}$ $\in$ Tensor(K, 4, 4), $P_{cam2ego}$ $\in$ Tensor(K, 4, 4), $P_{ego2global}$ $\in$ Tensor(K, 4, 4), $P_{intrinsics}$ $\in$ Tensor(K, 4, 4), voxel\_state $\in$ Tensor(H,W,Z), voxel\_label $\in$ Tensor(H,W,Z), pc\_range $\in$ List[6], voxel\_size $\in$ List[3],  spatial\_shape $\in$ List[3]}
\KwResult{$update\_voxel\_state$ $\in$ Tensor(H,W,Z)}
\SetKwProg{Fn}{Function}{:}{}
\Fn{calculate\_Camera\_visibility}{
    Initialize $origins\_list \gets List[]$, $uvs\_list \gets List[]$ \\
    $update\_voxel\_state \gets NOT\_OBSERVED$ \\
    \For{k in 0 to K}{
        /* Generate meshgrid points for image */ \\
        $uvs  \in Tensor(2, h * w) \gets meshgrid(Image[k])$ \\        
        $depth \gets Full((1, h * w), fill\_value=DEPTH\_MAX)$\\
        $uvs \gets concatenate([uvs, Ones((1, h * w))])$ \\
        $uvs \gets uvs * depth.repeat(3, 1)$ \\
        Convert uvs from Image to ego coordinate using $P_{cam2ego}$[k] and $P_{intrinsics}$[k]\\
        $uvs \gets uvs.transpose()$ \\
        $origin \gets Zeros((4, 4))$\\
        $origin[3, 3] \gets 1$\\
        Convert origin from Camera to ego coordinate using $P_{cam2ego}$[k]\\
        $origin \gets origin.reshape(1, -1).expand(uvs.shape[0], 3)$\;
        
        Add $uvs$ to $uvs\_list$\\
        Add $origin$ to $origins\_list$\\
    }
    $uv2points \gets concatenate(uvs\_list\_list)$ \\
    $origins \gets concatenate(origins\_list)$\\

    \For{i in 0 to N}{
        $ray\_start \gets origins[i]$\\
        $ray\_end \gets uv2points[i]$\\

        \For{voxel\_index in ray\_casting(ray\_start, ray\_end, pc\_range, voxel\_size, spatial\_shape)}{            
            \eIf{$voxel\_state == OCCUPIED$}{
                $update\_voxel\_state \gets OCCUPIED\;$
            }{
               \eIf{$voxel\_state == FREE$}{
                    $update\_voxel\_state \gets FREE\;$
               }{
                    $update\_voxel\_state \gets NOT\_OBSERVED\;$
               } 
            }
        }
    }
}
\caption{Camera Visibility}
\label{alg:Camera_visibility}
\end{algorithm}

\begin{figure*}[t]
    \centering
    \includegraphics[width=0.98\linewidth]{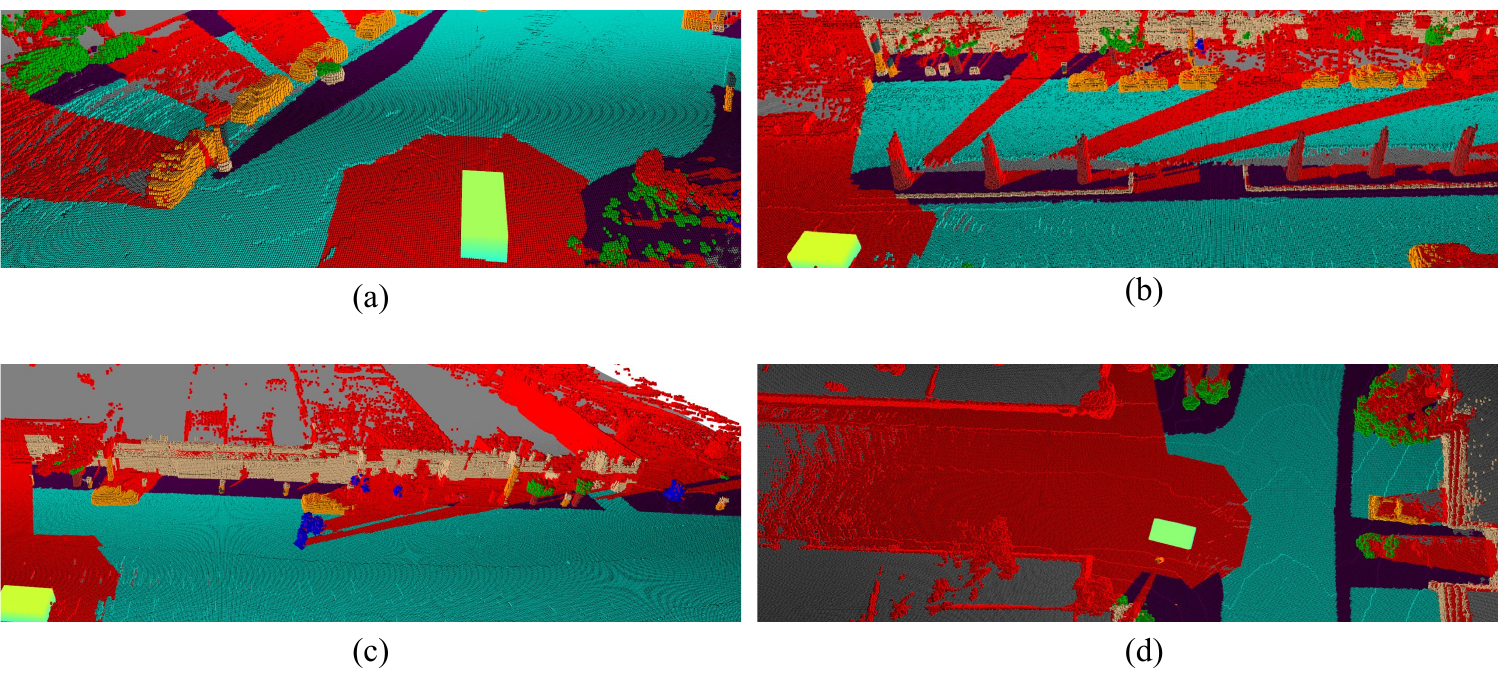}
    \caption{\textbf{Occlusion reasoning and camera visibility.} Grey voxels are unobserved in the LiDAR view and red voxels are observed in the accumulative LiDAR view but unobserved in the current camera view.
}
\label{fig:mask}
\end{figure*}

\paragraph{Visualization.} Accurately determining the visibility of a voxel is crucial for the 3D occupancy prediction task, as it helps eliminate training and evaluation ambiguity. As discussed in Section 4, Figure~\ref{fig:mask} illustrates the ``unobserved" voxels in the camera view due to occlusion. The yellow-green cube represents the ego vehicle, and the red-colored voxels are the ``unobserved" voxels determined by our visibility mask generation procedure. Figure~\ref{fig:mask}(a) shows the blind spots of ego vehicles and how parked vehicles at the roadside occlude the area behind them.  Figure.~\ref{fig:mask}(b) mainly shows that in the current camera views, the drivable surface and the buildings behind the tree trunks are occluded. In the right part of the image in Figure.~\ref{fig:mask}(c), voxels that represent buildings behind walls are marked as ``unobserved". As illustrated in Figure.~\ref{fig:mask}(d), the Waymo dataset doesn't provide the back-view camera image, leading to the blind spots in a certain range of angles behind the vehicle. By accurately determining voxel visibility, we can improve the accuracy and reliability of our 3D occupancy prediction model, which is critical for autonomous driving systems.

\section{3D-2D Consistency}
\begin{figure}[t]
    \centering
    \subfloat[CAMERA\_FRONT]{
    \includegraphics[width=0.96\textwidth]{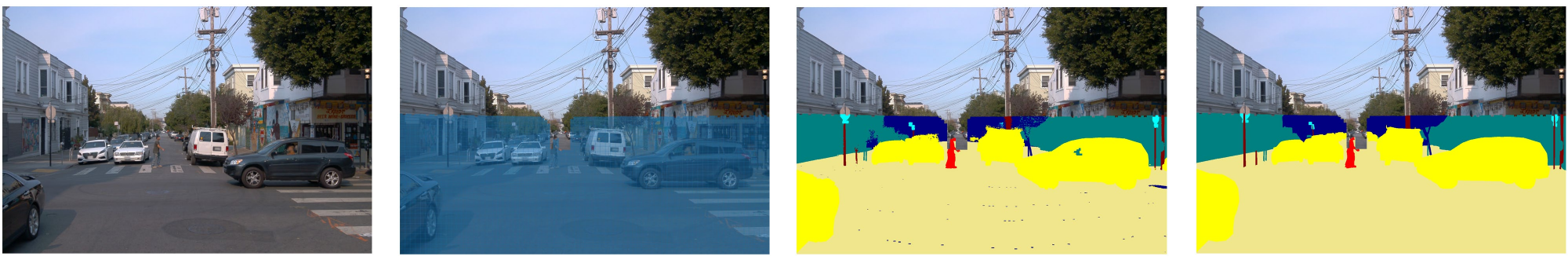}
    \label{fig:consis_0}
    }
    \quad
    \subfloat[CAMERA\_FRONT\_LEFT]{
    \includegraphics[width=0.96\textwidth]{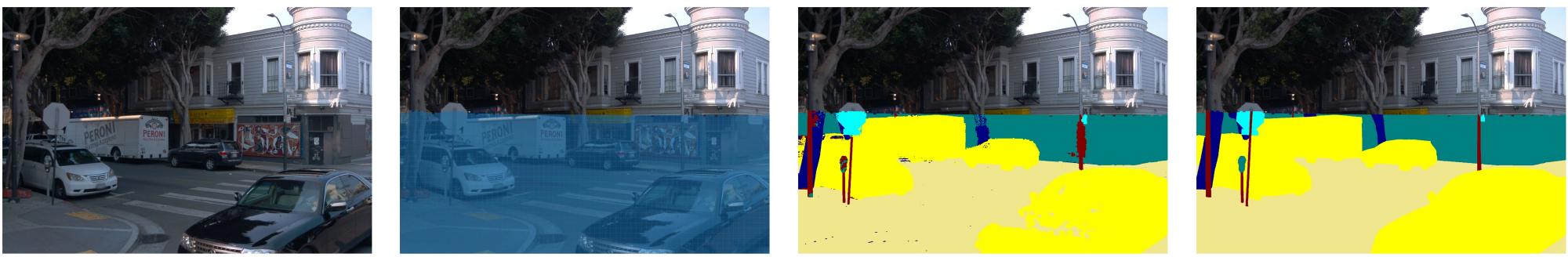}
    \label{fig:consis_1}
    }
    \quad
    \subfloat[CAMERA\_LEFT]{
    \includegraphics[width=0.96\textwidth]{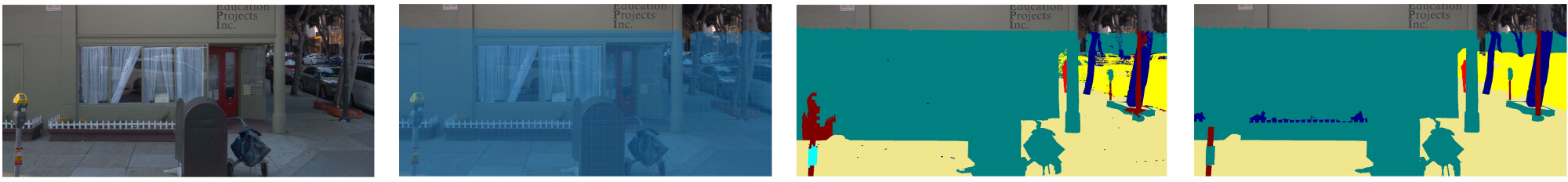}
    \label{fig:consis_2}
    }
    \quad
    \subfloat[CAMERA\_FRONT\_RIGHT]{
    \includegraphics[width=0.96\textwidth]{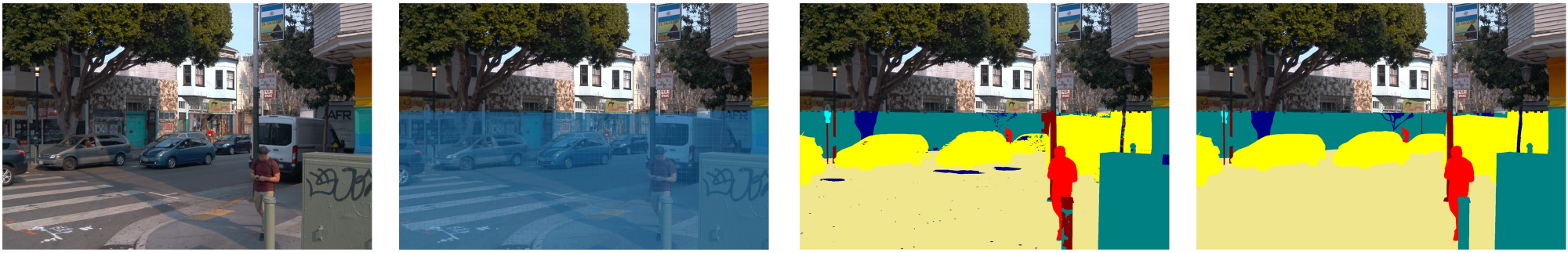}
    \label{fig:consis_3}
    }
    \quad
    \subfloat[CAMERA\_RIGHT]{
    \includegraphics[width=0.96\textwidth]{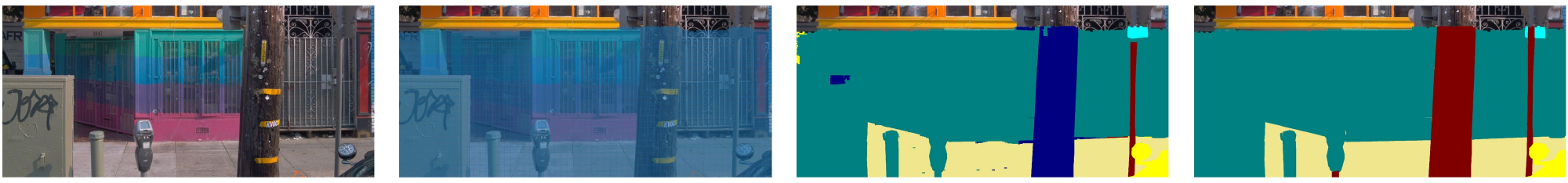}
    \label{fig:consis_4}
    }

    \caption{\textbf{Visualization of 3D-2D consistency.} From right to left are the visualization of original images, 2D ROI, 3D voxel semantics, 2D pixel semantics; From top to bottom are the results of CAMERA$\_$FRONT, CAMERA$\_$FRONT$\_$LEFT, CAMERA$\_$LEFT, CAMERA$\_$FRONT$\_$RIGHT, and CAMERA$\_$RIGHT.}
    \vspace{10pt}
    \label{fig:consis}
\end{figure}

Figure \ref{fig:consis} illustrates a visualization of the 3D-2D consistency evaluation conducted using the Waymo dataset. From right to left, the figure displays the original image, the 2D ROI, 3D voxel semantics, and 2D pixel semantics. Vertically, the figure presents the results in the order of CAMERA\_FRONT, CAMERA\_FRONT\_LEFT, CAMERA\_LEFT, CAMERA\_FRONT\_RIGHT, and CAMERA\_RIGHT. The result for CAMERA\_BACK is notably absent due to the original Waymo dataset not including images from rear-view cameras.

The visualization results demonstrate that the semantic labels for 3D voxels, generated via our auto-labeling method, align consistently with the manually annotated 2D semantic labels. This underscores the effectiveness of our proposed method. In the majority of instances, our proposed 3D-2D consistency calculation method provides an accurate measurement of this consistency. However, in certain situations, such as in Figure \ref{fig:consis_4} where the 2D semantic labels incorrectly annotated a tree trunk as a pole by humans, there can be a notable impact on the 3D-2D consistency metrics.

\vspace{15pt}
\section{Datasheet}

\begin{enumerate}

\item \textbf{For what purpose was the dataset created?} Was there a specific task in mind? Was there a specific gap that needed to be filled? Please provide a description.

\begin{itemize}
\item Occ3D was created as a benchmark for 3D Occupancy Prediction task. The goal of this task is to predict the 3D occupancy of the scene. Understanding the 3D surroundings including the background stuffs and foreground objects is important for autonomous driving. In the traditional 3D object detection task, a foreground object is represented by the 3D bounding box. However, the geometrical shape of the object is complex, which can not be represented by a simple 3D box, and the perception of the background stuffs is absent. The benchmark is a voxelized representation of the 3D space, and the occupancy state and semantics of the voxel in 3D space are jointly estimated in this task. The complexity of this task lies in the dense prediction of 3D space given the surround-view images.

\end{itemize}

\item \textbf{Who created the dataset (e.g., which team, research group) and on behalf of which entity (e.g., company, institution, organization)?}

\begin{itemize}
\item This dataset is presented by Tsinghua MARS Lab.
\end{itemize}

\item \textbf{Who funded the creation of the dataset?} If there is an associated grant, please provide the name of the grantor and the grant name and number.

\begin{itemize}
\item This work was sponsored by Tsinghua University.
\end{itemize}

\item \textbf{Any other comments?}

\begin{itemize}
\item No.
\end{itemize}

\subsection{Composition}

\item \textbf{What do the instances that comprise the dataset represent (e.g., documents, photos, people, countries)?} \textit{Are there multiple types of instances (e.g., movies, users, and ratings; people and interactions between them; nodes and edges)? Please provide a description.}

\begin{itemize}
\item 
We provide 40,000 samples for Occ3D-nuScenes and 200,000 samples for Occ3D-Waymo. Each sample in Occ3D-nuScenes consists of the following: 6 RGB images; 1 LiDAR point cloud; 1 3D voxel semantic ground-truth; 1 LiDAR visibility mask; 1 camera visibility mask; 1 metadata. Each sample in Occ3D-Waymo consists of the following: 5 RGB images; 1 LiDAR point cloud; 1 3D voxel semantic ground-truth; 1 LiDAR visibility mask; 1 camera visibility mask; 1 metadata.
We made our benchmark openly available on the Occ3D github page(\url{https://github.com/Tsinghua-MARS-Lab/Occ3D}).

\end{itemize}

\item \textbf{How many instances are there in total (of each type, if appropriate)?}

\begin{itemize}
\item For Occ3D-nuScenes, there are 600 scenes for training, 150 scenes for valuation, 250 scenes for testing, 40,000 frames in total. For Occ3D-Waymo, there are 798 scenes for training, 202 scenes for valuation, 150 scenes for testing, 200,000 frames in total. 

\end{itemize}

\item \textbf{Does the dataset contain all possible instances or is it a sample (not necessarily random) of instances from a larger set?} \textit{If the dataset is a sample, then what is the larger set? Is the sample representative of the larger set (e.g., geographic coverage)? If so, please describe how this representativeness was validated/verified. If it is not representative of the larger set, please describe why not (e.g., to cover a more diverse range of instances, because instances were withheld or unavailable).}

\begin{itemize}
\item Both nuScenes and Waymo are open-source datasets. We use the proposed auto-labeling method to derive Occ3D-nuScenes and Occ3D-Waymo. For Occ3D-nuScene, we use the annotated frames(2Hz) in nuScenes, which is representative; For Occ3D-Waymo, we use all samples of Waymo Open dataset.

\end{itemize}

\item \textbf{What data does each instance consist of?} \textit{``Raw'' data (e.g., unprocessed text or images) or features? In either case, please provide a description.}

\begin{itemize}
\item Each instance consist of RGB images, LiDAR point cloud, 3D voxel semantic ground-truth, LiDAR visibility mask, camera visibility mask and metadata.
\end{itemize}

\item \textbf{Is there a label or target associated with each instance?} \textit{If so, please provide a description.}

\begin{itemize}
\item There is a 3D voxel semantics label for each instance, which describe the semantic label of each voxel in the 3D scene.
\end{itemize}

\item \textbf{Is any information missing from individual instances?} \textit{If so, please provide a description, explaining why this information is missing (e.g., because it was unavailable). This does not include intentionally removed information, but might include, e.g., redacted text.}

\begin{itemize}
\item No.
\end{itemize}

\item \textbf{Are relationships between individual instances made explicit (e.g., users' movie ratings, social network links)?} \textit{If so, please describe how these relationships are made explicit.}

\begin{itemize}
\item No.
\end{itemize}

\item \textbf{Are there recommended data splits (e.g., training, development/validation, testing)?} \textit{If so, please provide a description of these splits, explaining the rationale behind them.}

\begin{itemize}
\item We use the original data splits in nuScenes and Waymo for Occ3D. For Occ3D-nuScenes, there are 600 train sequences, 150 validation sequences and 200 test sequences; For Occ3D-Waymo, there are 798 train sequences, 202 validation sequences and 150 test squences.
\end{itemize}

\item \textbf{Are there any errors, sources of noise, or redundancies in the dataset?} \textit{If so, please provide a description.}

\begin{itemize}
\item There exist noises in the dataset due to the LiDAR nosies and pose inaccuracies. 
\end{itemize}

\item \textbf{Is the dataset self-contained, or does it link to or otherwise rely on external resources (e.g., websites, tweets, other datasets)?} \textit{If it links to or relies on external resources, a) are there guarantees that they will exist, and remain constant, over time; b) are there official archival versions of the complete dataset (i.e., including the external resources as they existed at the time the dataset was created); c) are there any restrictions (e.g., licenses, fees) associated with any of the external resources that might apply to a future user? Please provide descriptions of all external resources and any restrictions associated with them, as well as links or other access points, as appropriate.} \\
\begin{itemize}
\item We release the Occ3D dataset on our GitHub repository: \url{https://github.com/Tsinghua-MARS-Lab/Occ3D}. More specifically, please use the following links
to visit the documentations and download instructions: \href{https://tsinghua-mars-lab.github.io/Occ3D/}{Occ3D-Webpage}. Our dataset is developed based on existing automonous driving dataset \href{https://nuscenes.org}{nuScenes} and \href{https://waymo.com/open/}{Waymo}
\end{itemize}



\item \textbf{Does the dataset contain data that might be considered confidential (e.g., data that is protected by legal privilege or by doctor–patient confidentiality, data that includes the content of individuals’ non-public communications)?} \textit{If so, please provide a description.}

\begin{itemize}
\item Our dataset is developed based on \href{https://nuscenes.org}{nuScenes}(developed by \href{https://motional.com}{Motional} )and \href{https://waymo.com/open/}{Waymo} (developed by \href{https://waymo.com/}{Waymo} ), which has already removed confidential data.
\end{itemize}

\item \textbf{Does the dataset contain data that, if viewed directly, might be offensive, insulting, threatening, or might otherwise cause anxiety?} \textit{If so, please describe why.}

\begin{itemize}
\item No.
\end{itemize}

\item \textbf{Does the dataset relate to people?} \textit{If not, you may skip the remaining questions in this section.}

\begin{itemize}
\item No.
\end{itemize}

\item \textbf{Does the dataset identify any subpopulations (e.g., by age, gender)?}

\begin{itemize}
\item No.
\end{itemize}

\item \textbf{Is it possible to identify individuals (i.e., one or more natural persons), either directly or indirectly (i.e., in combination with other data) from the dataset?} \textit{If so, please describe how.}

\begin{itemize}
\item No.
\end{itemize}

\item \textbf{Does the dataset contain data that might be considered sensitive in any way (e.g., data that reveals racial or ethnic origins, sexual orientations, religious beliefs, political opinions or union memberships, or locations; financial or health data; biometric or genetic data; forms of government identification, such as social security numbers; criminal history)?} \textit{If so, please provide a description.}

\begin{itemize}
\item No.
\end{itemize}

\item \textbf{Any other comments?}

\begin{itemize}
\item No. 
\end{itemize}

\subsection{Collection Process}

\item \textbf{How was the data associated with each instance acquired?} \textit{Was the data directly observable (e.g., raw text, movie ratings), reported by subjects (e.g., survey responses), or indirectly inferred/derived from other data (e.g., part-of-speech tags, model-based guesses for age or language)? If data was reported by subjects or indirectly inferred/derived from other data, was the data validated/verified? If so, please describe how.}

\begin{itemize}
\item Our data is developing based on published data \href{https://nuscenes.org}{nuScenes} and \href{https://waymo.com/open/}{Waymo} using a designed auto-labeling method mentioned before. 
\end{itemize}

\item \textbf{What mechanisms or procedures were used to collect the data (e.g., hardware apparatus or sensor, manual human curation, software program, software API)?} \textit{How were these mechanisms or procedures validated?}

\begin{itemize}
\item We ran a auto-labeling script in python to generate the ground-truth labels. We use hundred of small CPU nodes, and few GPU nodes. They were validated by manual inspection of the results and 2D-3D consistency quality check we described in the body part.
\end{itemize}

\item \textbf{If the dataset is a sample from a larger set, what was the sampling strategy (e.g., deterministic, probabilistic with specific sampling probabilities)?}

\begin{itemize}
\item We use full-set provided by \href{https://nuscenes.org}{nuScenes} and \href{https://waymo.com/open/}{Waymo}.

\end{itemize}

\item \textbf{Who was involved in the data collection process (e.g., students, crowdworkers, contractors) and how were they compensated (e.g., how much were crowdworkers paid)?}

\begin{itemize}
\item No crowdworkers were involved in the curation of the dataset. Open-source researchers and developers enabled its creation for no payment.
\end{itemize}

\item \textbf{Over what timeframe was the data collected? Does this timeframe match the creation timeframe of the data associated with the instances (e.g., recent crawl of old news articles)?} \textit{If not, please describe the timeframe in which the data associated with the instances was created.}

\begin{itemize}
\item The 3D occupancy ground-truth data was generated in 2023, while the source sensor data was created in 2019 for nuScenes and 2020 for Waymo.
\end{itemize}

\item \textbf{Were any ethical review processes conducted (e.g., by an institutional review board)?} \textit{If so, please provide a description of these review processes, including the outcomes, as well as a link or other access point to any supporting documentation.}

\begin{itemize}
\item The source sensor data for nuScenes and Waymo had been conducted ethical review processes by Motional and Waymo, which can be referred to \href{https://nuscenes.org}{nuScenes} and \href{https://waymo.com/open/}{Waymo}, respectively.
\end{itemize}

\item \textbf{Did you collect the data from the individuals in question directly, or obtain it via third parties or other sources (e.g., websites)?}

\begin{itemize}
\item We retrieve the data from the open source datasets \href{https://nuscenes.org}{nuScenes} and \href{https://waymo.com/open/}{Waymo}.

\end{itemize}

\item \textbf{Were the individuals in question notified about the data collection?} \textit{If so, please describe (or show with screenshots or other information) how notice was provided, and provide a link or other access point to, or otherwise reproduce, the exact language of the notification itself.}

\begin{itemize}
\item The Occ3D dataset is developed based on open-source dataset and following the open-source license.
\end{itemize}

\item \textbf{Did the individuals in question consent to the collection and use of their data?} \textit{If so, please describe (or show with screenshots or other information) how consent was requested and provided, and provide a link or other access point to, or otherwise reproduce, the exact language to which the individuals consented.}

\begin{itemize}

\item The Occ3D dataset is developed on open-source dataset and obey the license.
\end{itemize}

\item \textbf{If consent was obtained, were the consenting individuals provided with a mechanism to revoke their consent in the future or for certain uses?} \textit{If so, please provide a description, as well as a link or other access point to the mechanism (if appropriate).}

\begin{itemize}


\item Users have a possibility to check for the presence of the links in our dataset leading to their data on public internet by using the search tool provided by Occ3D, accessible at \href{https://tsinghua-mars-lab.github.io/Occ3D/}{Occ3D-Webpage}. If users wish to revoke their consent after finding sensitive data, they can contact the hosting party and request to delete the content from the underlying website. Please leave the message in \href{https://github.com/Tsinghua-MARS-Lab/Occ3D/issues}{GitHub Issue} to request removal of the links from the dataset. 
\end{itemize}

\item \textbf{Has an analysis of the potential impact of the dataset and its use on data subjects (e.g., a data protection impact analysis) been conducted?} \textit{If so, please provide a description of this analysis, including the outcomes, as well as a link or other access point to any supporting documentation.}

\begin{itemize}


\item We develop our dataset based on open source dataset \href{https://nuscenes.org}{nuScenes} and \href{https://waymo.com/open/}{Waymo} publised by \href{https://motional.com}{Motional} and \href{https://waymo.com}{Waymo}. The published dataset has been seriously considered of it's potential impact and its use on data subjects.

\end{itemize}

\item \textbf{Any other comments?}

\begin{itemize}
\item No.
\end{itemize}

\subsection{Preprocessing, Cleaning, and/or Labeling}

\item \textbf{Was any preprocessing/cleaning/labeling of the data done (e.g., discretization or bucketing, tokenization, part-of-speech tagging, SIFT feature extraction, removal of instances, processing of missing values)?} \textit{If so, please provide a description. If not, you may skip the remainder of the questions in this section.}

\begin{itemize}

\item We use an auto-labeling preprocessing script to generate the 3D voxel semantic labels of the dataset. Beside this, no preprocessing or labelling is done.

\end{itemize}

\item \textbf{Was the ``raw'' data saved in addition to the preprocessed/cleaned/labeled data (e.g., to support unanticipated future uses)?} \textit{If so, please provide a link or other access point to the ``raw'' data.}

\begin{itemize}
\item Yes, we provide the original open source dataset and the auto-labeled Occ3D dataset.
\end{itemize}

\item \textbf{Is the software used to preprocess/clean/label the instances available?} \textit{If so, please provide a link or other access point.}

\begin{itemize}
\item No.

\end{itemize}

\item \textbf{Any other comments?}

\begin{itemize}
\item No.
\end{itemize}

\subsection{Uses}

\item \textbf{Has the dataset been used for any tasks already?} \textit{If so, please provide a description.}

\begin{itemize}
\item No.
\end{itemize}

\item \textbf{Is there a repository that links to any or all papers or systems that use the dataset?} \textit{If so, please provide a link or other access point.}

\begin{itemize}
\item No.
\end{itemize}

\item \textbf{What (other) tasks could the dataset be used for?}

\begin{itemize}
\item We encourage future researchers to curate Occ3D for several tasks. For instance, we hope that researchers can use the Occ3D we provide to study how to better promote some downstream tasks such as autonomous driving prediction and planning.
\end{itemize}

\item \textbf{Is there anything about the composition of the dataset or the way it was collected and preprocessed/cleaned/labeled that might impact future uses?} \textit{For example, is there anything that a future user might need to know to avoid uses that could result in unfair treatment of individuals or groups (e.g., stereotyping, quality of service issues) or other undesirable harms (e.g., financial harms, legal risks) If so, please provide a description. Is there anything a future user could do to mitigate these undesirable harms?}

\begin{itemize}

\item No.
\end{itemize}

\item \textbf{Are there tasks for which the dataset should not be used?} \textit{If so, please provide a description.}

\begin{itemize}
\item Due to the known biases of the dataset, under no circumstance should any models be put into production using the dataset as is. It is neither safe nor responsible. As it stands, the dataset should be solely used for research purposes in its uncurated state.
\end{itemize}

\item \textbf{Any other comments?}

\begin{itemize}
\item No.
\end{itemize}

\subsection{Distribution}

\item \textbf{Will the dataset be distributed to third parties outside of the entity (e.g., company, institution, organization) on behalf of which the dataset was created?} \textit{If so, please provide a description.}

\begin{itemize}
\item Yes, the dataset will be open-source.
\end{itemize}

\item \textbf{How will the dataset be distributed (e.g., tarball on website, API, GitHub)?} \textit{Does the dataset have a digital object identifier (DOI)?}

\begin{itemize}

\item The data is available through \url{https://github.com/Tsinghua-MARS-Lab/Occ3D}.
\end{itemize}

\item \textbf{When will the dataset be distributed?}

\begin{itemize}
\item 31/03/2023 and onward.
\end{itemize}

\item \textbf{Will the dataset be distributed under a copyright or other intellectual property (IP) license, and/or under applicable terms of use (ToU)?} \textit{If so, please describe this license and/or ToU, and provide a link or other access point to, or otherwise reproduce, any relevant licensing terms or ToU, as well as any fees associated with these restrictions.}

\begin{itemize}
\item The Occ3D dataset is published under \href{https://en.wikipedia.org/wiki/MIT_License}{MIT} license, which means everyone can use this dataset for non-commercial research purpose. The original nuScenes dataset is released under the \href{https://creativecommons.org/licenses/by-nc-sa/4.0/legalcode}{CC BY-NC-SA 4.0}. The original Waymo dataset is released under the \href{https://waymo.com/open/terms/}{Waymo Dataset License Agreement for Non-Commercial Use (August 2019)} License. 
\end{itemize}

\item \textbf{Have any third parties imposed IP-based or other restrictions on the data associated with the instances?} \textit{If so, please describe these restrictions, and provide a link or other access point to, or otherwise reproduce, any relevant licensing terms, as well as any fees associated with these restrictions.}

\begin{itemize}
\item The original nuScenes dataset is released under the \href{https://creativecommons.org/licenses/by-nc-sa/4.0/legalcode}{CC BY-NC-SA 4.0}, and the for the restrictions, please refer to \href{https://nuscenes.org}{nuScenes}. The original Waymo dataset is released under the \href{https://waymo.com/open/terms/}{Waymo Dataset License Agreement for Non-Commercial Use (August 2019)} License, and the for the restrictions, please refer to \href{https://waymo.com/open/}{Waymo}.  
\end{itemize}

\item \textbf{Do any export controls or other regulatory restrictions apply to the dataset or to individual instances?} \textit{If so, please describe these restrictions, and provide a link or other access point to, or otherwise reproduce, any supporting documentation.}

\begin{itemize}
\item No.
\end{itemize}

\item \textbf{Any other comments?}

\begin{itemize}
\item No.
\end{itemize}

\subsection{Maintenance}

\item \textbf{Who will be supporting/hosting/maintaining the dataset?}

\begin{itemize}
\item Tsinghua MARS Lab will support hosting of the dataset.
\end{itemize}

\item \textbf{How can the owner/curator/manager of the dataset be contacted (e.g., email address)?}

\begin{itemize}
\item \url{https://github.com/Tsinghua-MARS-Lab/Occ3D/issues}
\end{itemize}

\item \textbf{Is there an erratum?} \textit{If so, please provide a link or other access point.}

\begin{itemize}
\item There is no erratum for our initial release. Errata will be documented as future releases on the dataset website.
\end{itemize}

\item \textbf{Will the dataset be updated (e.g., to correct labeling errors, add new instances, delete instances)?} \textit{If so, please describe how often, by whom, and how updates will be communicated to users (e.g., mailing list, GitHub)?}

\begin{itemize}
\item We will continue to support Occ3D dataset.
\end{itemize}

\item \textbf{If the dataset relates to people, are there applicable limits on the retention of the data associated with the instances (e.g., were individuals in question told that their data would be retained for a fixed period of time and then deleted)?} \textit{If so, please describe these limits and explain how they will be enforced.}

\begin{itemize}
\item No.
\end{itemize}

\item \textbf{Will older versions of the dataset continue to be supported/hosted/maintained?} \textit{If so, please describe how. If not, please describe how its obsolescence will be communicated to users.}

\begin{itemize}
\item Yes. We will continue to support Occ3D dataset in \href{https://tsinghua-mars-lab.github.io/Occ3D/}{our github page}.
\end{itemize}

\item \textbf{If others want to extend/augment/build on/contribute to the dataset, is there a mechanism for them to do so?} \textit{If so, please provide a description. Will these contributions be validated/verified? If so, please describe how. If not, why not? Is there a process for communicating/distributing these contributions to other users? If so, please provide a description.}

\begin{itemize}
\item Yes, they can driectly developing on open scource dataset \href{https://nuscenes.org}{nuScenes} and \href{https://waymo.com/open/}{Waymo} dataset or concat us via \href{https://github.com/Tsinghua-MARS-Lab/Occ3D/issues}{GitHub Issue}.
\end{itemize}

\item \textbf{Any other comments?}

\begin{itemize}
\item No.
\end{itemize}

\end{enumerate}
\section*{Checklist}
\begin{enumerate}

\item For all authors...
\begin{enumerate}
  \item Do the main claims made in the abstract and introduction accurately reflect the paper's contributions and scope?
    \answerYes{}
  \item Did you describe the limitations of your work?
    \answerYes{See Section~\ref{sec:conclusion}.}
  \item Did you discuss any potential negative societal impacts of your work?
    \answerYes{}
  \item Have you read the ethics review guidelines and ensured that your paper conforms to them?
    \answerYes{Our dataset cannot cover all objects in the real world. There may be safety risks when applying algorithms developed based on this dataset to real roads.}
\end{enumerate}

\item If you are including theoretical results...
\begin{enumerate}
  \item Did you state the full set of assumptions of all theoretical results?
    \answerNA{}
	\item Did you include complete proofs of all theoretical results?
    \answerNA{}
\end{enumerate}

\item If you ran experiments (e.g. for benchmarks)...
\begin{enumerate}
  \item Did you include the code, data, and instructions needed to reproduce the main experimental results (either in the supplemental material or as a URL)?
    \answerYes{See Abstract.}
  \item Did you specify all the training details (e.g., data splits, hyperparameters, how they were chosen)?
    \answerYes{See the Appendix.}
	\item Did you report error bars (e.g., with respect to the random seed after running experiments multiple times)?
    \answerNo{}We did not repeat experiments multiple times.
	\item Did you include the total amount of compute and the type of resources used (e.g., type of GPUs, internal cluster, or cloud provider)?
    \answerYes{See the Appendix.}
\end{enumerate}

\item If you are using existing assets (e.g., code, data, models) or curating/releasing new assets...
\begin{enumerate}
  \item If your work uses existing assets, did you cite the creators?
    \answerYes{}
  \item Did you mention the license of the assets?
    \answerYes{The nuScenes dataset is released under the \href{https://creativecommons.org/licenses/by-nc-sa/4.0/legalcode}{CC BY-NC-SA 4.0}. The Waymo dataset is released under the \href{https://waymo.com/open/terms/}{Waymo Dataset License Agreement for Non-Commercial Use (August 2019)} License.}
  \item Did you include any new assets either in the supplemental material or as a URL?
    \answerYes{See Abstract.}
  \item Did you discuss whether and how consent was obtained from people whose data you're using/curating?
    \answerNo{The assets used are public.}
  \item Did you discuss whether the data you are using/curating contains personally identifiable information or offensive content?
    \answerNA{}
\end{enumerate}

\item If you used crowdsourcing or conducted research with human subjects...
\begin{enumerate}
  \item Did you include the full text of instructions given to participants and screenshots, if applicable?
    \answerNA{}
  \item Did you describe any potential participant risks, with links to Institutional Review Board (IRB) approvals, if applicable?
    \answerNA{}
  \item Did you include the estimated hourly wage paid to participants and the total amount spent on participant compensation?
    \answerNA{}
\end{enumerate}

\end{enumerate}

\end{document}